\documentclass{article}

\usepackage{arxiv}
\usepackage[utf8]{inputenc} 
\usepackage[T1]{fontenc}    
\usepackage{hyperref}       
\usepackage{url}            
\usepackage{booktabs}       
\usepackage{amsfonts}       
\usepackage{nicefrac}       
\usepackage{microtype}      
\usepackage{graphicx}
\usepackage{doi}
\usepackage{pgfplots}
\usepackage{booktabs}
\usepackage{amsmath}
\usepackage{multirow}
\usepackage{amssymb}
\usepackage{mathrsfs}
\usepackage{wasysym}
\usepackage{nth}
\usepackage{xspace}

\makeatletter
\providecommand*{\input@path}{}
\g@addto@macro\input@path{{./figures/}{./tables/}{./contents/}}
\makeatother

\usepackage[capitalize, nameinlink]{cleveref}
\crefname{section}{Sec.}{Secs.}
\Crefname{section}{Sec.}{Secs.}
\crefname{subsection}{Sec.}{Secs.}
\Crefname{subsection}{Sec.}{Secs.}
\Crefname{table}{Table}{Tables}
\crefname{table}{Table}{Tables}
\Crefname{figure}{Fig.}{Figs.}
\crefname{figure}{Fig.}{Figs.}
\Crefname{equation}{Equation}{Equations}
\crefname{equation}{Equation}{Equations}

\newcommand{\Fup}[1]{\text{F}_{\text{up}\text{,#1}}}
\newcommand{\Fdown}[1]{\text{F}_{\text{down}\text{,#1}}}
\newcommand{\M}{\mathscr{M}}
\newcommand{\dist}{\text{d}} 

\newcommand{\Diff}[1]{\boldsymbol{D}_{\text{#1}}}
\newcommand{\px}{\,\text{px}}
\newcommand{\loss}[1]{\mathscr{L}_{\text{#1}}}
\newcommand{\im}[1]{\boldsymbol{I}_{\text{#1}}} 
\newcommand{\norm}[1]{\left\lVert#1\right\rVert}

\newlength\figureheight
\newlength\figurewidth

\makeatletter
\DeclareRobustCommand\onedot{\futurelet\@let@token\@onedot}
\def\@onedot{\ifx\@let@token.\else.\null\fi\xspace}
 
\def\ie{\emph{i.e}\onedot} 
\def\cf{\emph{cf}\onedot}

\def\etal{\emph{et al}\onedot}

\def\BT{\emph{BT}\xspace}
\def\ST{\emph{ST}\xspace}
\makeatother

\title{Susceptibility to Image Resolution in Face Recognition
and Training Strategies to Enhance Robustness}

\author{\href{https://orcid.org/0000-0002-0503-4600}{\includegraphics[scale=0.06]{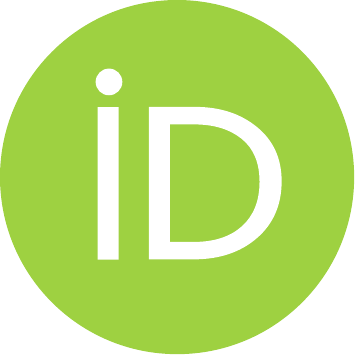}\hspace{1mm}Martin~Knoche}\quad Stefan~H\"orman\quad\href{https://orcid.org/0000-0003-1096-1596}{\includegraphics[scale=0.06]{orcid.pdf}\hspace{1mm}Gerhard ~Rigoll}\\
	Chair of Human-Machine Communication\\
	Technical University\\
	Munich, Germany \\
	\texttt{\href{Martin.Knoche@tum.de}{Martin.Knoche@tum.de}} \\
}

\renewcommand{\headeright}{A Preprint}
\renewcommand{\undertitle}{A Preprint \\ 
Published in Leibnitz Transactions on Embedded Systems 2022 \\ 
DOI: \url{https://doi.org/10.4230/LITES.8.1.1}}
\renewcommand{\shorttitle}{\textit{M. Knoche et al.} Susceptibility to Image Resolution in Face Recognition}

\hypersetup{
pdftitle={btm-stm},
pdfsubject={},
pdfauthor={Martin~Knoche, Stefan~H\"orman, Gerhard~Rigoll},
pdfkeywords={recognition, resolution, cross, face, identification, verification},
}

\begin{document}
\maketitle

\begin{abstract}
	Face recognition approaches often rely on equal image resolution for verification faces on two images. However, in practical applications, those image resolutions are usually not in the same range due to different image capture mechanisms or sources. In this work, we first analyze the impact of image resolutions on the face verification performance with a state-of-the-art face recognition model. For images, synthetically reduced to $5\,\times\,5\px$ resolution, the verification performance drops from $99.23\%$ increasingly down to almost $55\%$. Especially, for cross-resolution image pairs (one high- and one low-resolution image), the verification accuracy decreases even further. We investigate this behavior more in-depth by looking at the feature distances for every 2-image test pair. To tackle this problem, we propose the following two methods: 1) Train a state-of-the-art face-recognition model straightforward with $50\%$ low-resolution images directly within each batch. 2) Train a siamese-network structure and adding a cosine distance feature loss between high- and low-resolution features. Both methods show an improvement for cross-resolution scenarios and can increase the accuracy at very low resolution to approximately $70\%$. However, a disadvantage is that a specific model needs to be trained for every resolution-pair. Thus, we extend the aforementioned methods by training them with multiple image resolutions at once. The performances for particular testing image resolutions are slightly worse, but the advantage is that this model can be applied to arbitrary resolution images and achieves overall a better performance ($97.72\%$ compared to $96.86\%$). Due to the lack of a benchmark for arbitrary resolution images for the cross-resolution and equal-resolution task, we propose an evaluation protocol for five well-known datasets, focusing on high, mid, and low-resolution images. \footnote{This work was partially supported by the \textit{Bayerische Staatsministerium f\"ur Wirtschaft, Energie und Technologie}
within the framework of a funding program of Informations- und Kommunikationstechnik for the project
"Grundrechtskonforme Gesichtserkennung im \"offentlichen Raum" (e-freedom).}
\end{abstract}

\section{Introduction}
\label{sec:introduction}
In recent years, face recognition has gained progressively more attraction. Due to the availability of powerful GPUs and novel datasets with up to $87\mathrm{k}$ identities, e.g. Microsoft's Celeb Dataset (MS1M)~\cite{guo2016ms}, research shifted more and more to deep-learning-based approaches. Nowadays, methods are generally based on the same principle as the method of choice: Representing the face using a deep convolutional neural network (DCNN) embedding. DCNNs are trained to map a facial image, typically after a pose alignment step, into a feature space such that intra-class distances are minimized and inter-class distances are maximized. One of the first approaches was proposed by Szegedy \etal~\cite{szegedy2015going} in 2014 with a 9-layer convolutional neural network. 

Typically, face recognition networks learn a projection into a distinct feature space according to their training dataset. Popular approaches are trained with datasets containing sufficient amount of high-resolution (HR) images. However, in real-world applications, the image quality is often inferior. Not only illumination conditions are different, also the resolution of a captured face can be of arbitrary size. For example, surveillance cameras capturing faces at very low resolutions, in contrast to very high-quality mug-shots-like passport images. Moreover, photos uploaded on social media platforms might show people in the background, hence, only tiny faces occur, which should also be recognized automatically by an algorithm. Current state-of-the-art approaches are susceptible against resolution and do not solve this problem satisfactorily. We analyze this in \cref{sec:analysis} more detailed. 
Many works~\cite{li2019low,cheng2018lowres,aghdam2019exploring} address low-resolution (LR) images for face recognition. However, they are using the same resolution for all test images. In real-world scenarios, image resolutions are likely different in the test setting. For example in social media networks people have a high resolution profile picture and should be recognized on photos, standing in the background, far away from the camera. Cross resolution (CR) face recognition is addressing the problem of comparing images with varying resolutions of images, which has yet found less attraction in the research. According to~\cite{singh2018magnifyme}, existing approaches can be grouped into two methods: 1) Transformation-based approaches, on the one hand, aim to transform LR images to HR images via identity preserving super-resolution algorithms~\cite{zhang2018super,dogan2019exemplar,nm2019identity,hsu2019sigan}. In the opposite image resolution case, they perform simple down-sampling from HR to LR images. On the other hand, algorithms try to project extracted LR features into a HR feature space or vice-versa. 2) Non-transformation based approaches~\cite{massoli2020cross,zeng2016towards} target to directly extract scale-invariant features into a common feature space.

\begin{figure}[t]
	\centering
	\includegraphics[width=\textwidth]{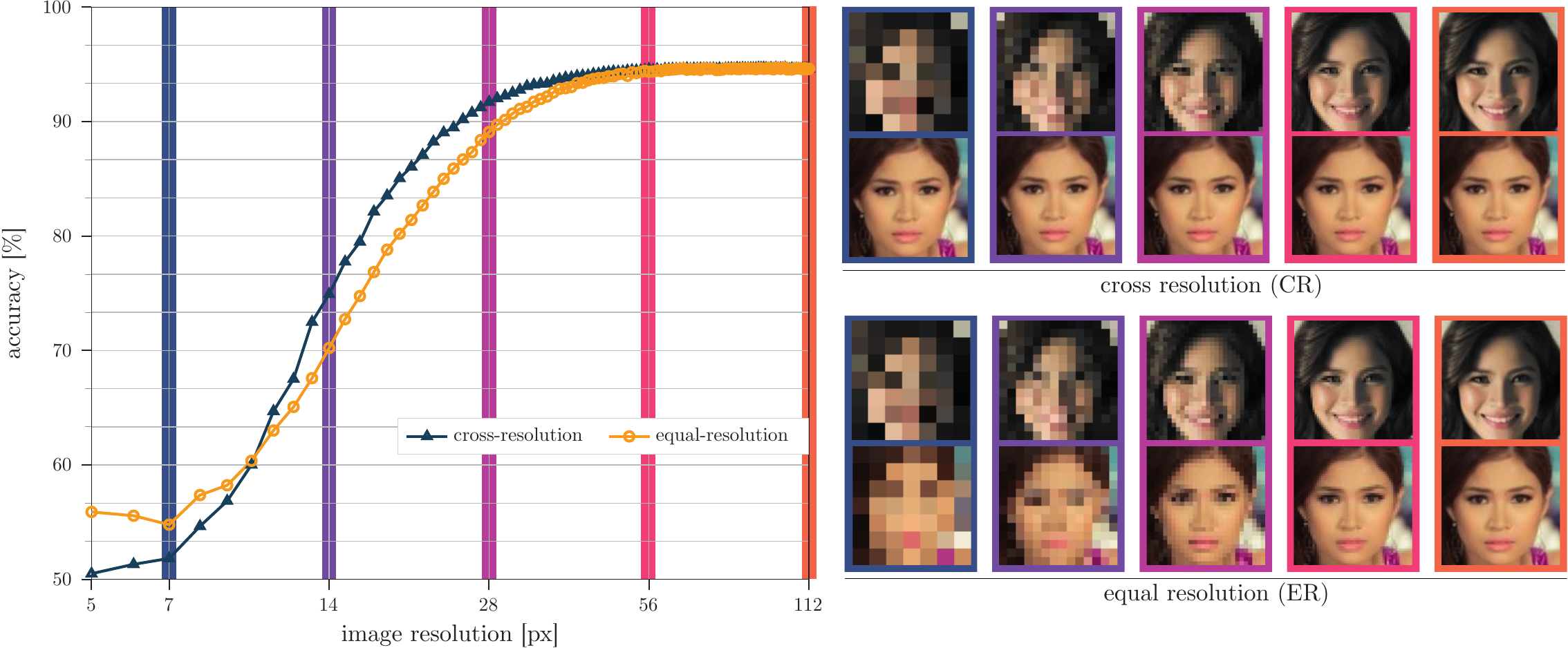}
	\caption{Average face verification accuracy across five popular datasets (LFW~\cite{lfw_dataset}, AgeBDB~\cite{moschoglou2017agedb}, CFP-FP~\cite{sengupta2016frontal}, CALFW~\cite{zheng2017cross} and CPLFW~\cite{zheng2018cross}) for cross-resolution and equal-resolution (left). Example image pairs are selected for five different resolutions (right).}
	\label{fig:avg_acc_vis}
\end{figure}

In this work, we first investigate the behavior of a state-of-the-art face recognition network~\cite{deng2019arcface} on different resolution images for the verification task. We differentiate between a CR and LR verification scenarios in our analysis (\cf \cref{fig:avg_acc_vis}). \Cref{fig:avg_acc_vis} shows that the average performance (recognition accuracy) across several datasets strongly drops for CR and LR scenarios. At resolutions below $10\,\times\,10\px$ the accuracy is slightly above $50\%$, which is only barely about guessing. We are not concentrating at image resolutions lower than $5\,\times\,5\px$. From this finding, we conclude that current state-of-the-art is susceptible against resolution and there is a huge gap for improvements, especially for very low image resolutions. 

Other works addressing the CR problem are often dealing with exactly two resolutions in training and testing. However, this is not sufficiently transferable to real-world applications. Zeng \etal~\cite{zeng2016towards} use a mix of two and a mix of four resolutions, but they report only results in face identification scenarios. Therefore, we additionally experimented with multiple image resolutions for training. In general, we distinguish between two-resolution and multi-resolution training, which means on the one hand, precisely two different resolutions for training and on the other hand, numerous resolutions during training, respectively. 
\newpage
In summary, our main contributions are:
\begin{itemize}
	\item We analyze the susceptibility for different image-resolution on face verification in-depth.
	\item We propose two intuitive, straightforward approaches and show performance improvements, especially for very low image resolutions.
	\item Moreover, we perform multi-resolution learning and show performance across several datasets.
	\item Lastly, we propose and publish three evaluation protocols focusing on low, mid and high resolution to measure the performance of multiple resolutions in the cross-resolution verification scenario. This is to our best knowledge the first benchmark for CR. 
\end{itemize}
\section{Related Work}
\label{sec:related_work}

\subsection{Generic Face Recognition}
Most face recognition methods in recent years focused on the networks loss to improve performance: FaceNet~\cite{schroff2015facenet} proposed triplet loss in order to maximize the distances between the anchor image and it's positive sample. SphereFace~\cite{liu2017sphereface} introduced angular softmax loss with a multiplicative angular margin, whereas CosFace~\cite{wang2018cosface} proposed an additive cosine margin. Finally, the authors of ArcFace~\cite{deng2019arcface} apply an Additive Angular Margin Loss function, which can effectively extent the discriminative power of features. Recently, Kim \etal~\cite{kim2020groupface} presented with GroupFace a novel architecture that utilizes multiple group-aware representations, to improve the quality of the feature. Wang \etal~\cite{wang2020hierarchical} proposed a hierarchical pyramid diverse attention network. The latter two methods clearly outperform the previously mentioned algorithms and become state-of-the-art. 

\subsection{Image Resolutions}
To the best of our knowledge, there exists no dataset containing the same real-world face images in different resolutions for a comparatively large number of identities like the MS1M dataset. For analyzing the impact of resolutions, it is crucial to have the same photo in different resolutions. Naturally, a picture is taken in exactly one resolution. Hence there are no large datasets for this case available. This requires a synthetic downsample method, which can be applied onto arbitrary datasets. According to Zhou and Süsstrunk~\cite{zhou2019kernel} a mapping from LR to HR images is often learned by synthetically downsampled HR images in order to retrieve training data for super-resolution approaches. They further state that the frequently used bicubic interpolation~\cite{keys1981cubic} significantly differs from real-world camera-blur. 

\subsection{Cross Resolution Face Recognition}
\begin{figure}[t]
	\centering
	\includegraphics[width=\textwidth]{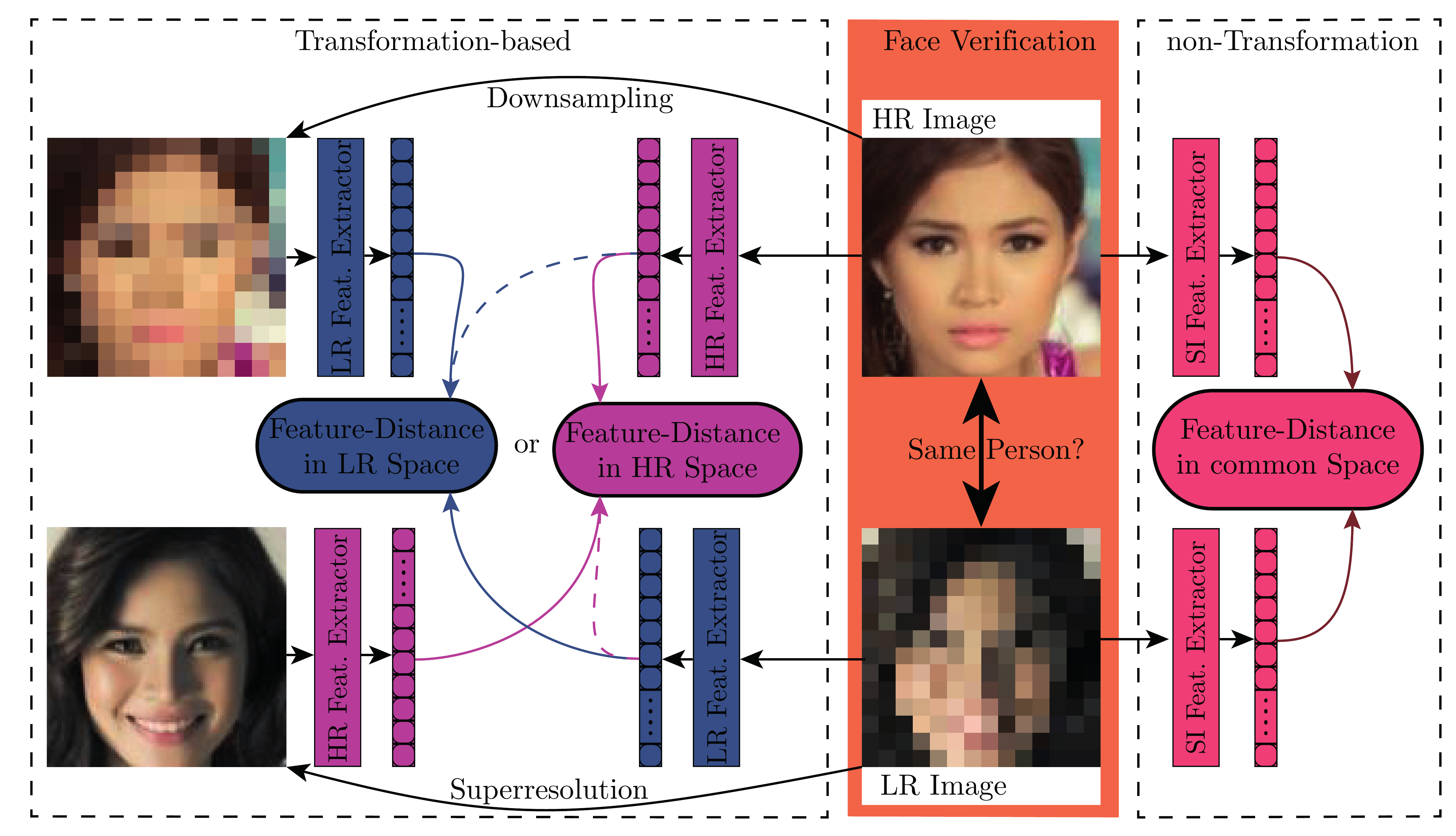}
	\caption{Methods in face recognition with CR images can be categorized into two folds: Non-transformation-based (right) approaches aim to learn directly scale-invariant image features. Transformation-based approaches (left) try to project learned image features or the images themselves to a common space.}
	\label{fig:overview_fr}
\end{figure}

As stated in \cref{sec:introduction}, face recognition in the context of CR can be categorized into two groups: Transformation-based and non-transformation-based methods. Wang \etal~\cite{wang2014low} show an exhaustive review of those methods for addressing cross-resolution face-recognition. \Cref{fig:overview_fr} gives a brief functional overview of those two methods. 

\subsubsection*{Transformation-based Methods}
These methods aim to focus on mapping images or extracted image features into a common feature space. 

Lu \etal~\cite{lu2018deep} presented a deep coupled ResNet model, containing one trunk network and two branch networks. The trunk network extracts features and the two branches networks transform HR and the corresponding LR features to a space where their difference is minimized. 

Zangeneh \etal~\cite{zangeneh2020low} proposed a two branch DCNN. While the LR branch consists of a super-resolution network combined with a feature-extraction network, the HR branch is only a feature-extraction network. Both branches are trained in three different training phases. For testing, images are fed through the branches depending on their resolution. A similar approach was used in \cite{khazaie2020ipu}, in which they trained a U-Net with a combination of reconstruction and identity preserving loss in order to super-resolve multi-scale low-resolution images. For feature extraction, they utilized a pretrained Inception-ResNet. 

The authors of~\cite{talreja2019attribute} proposed a coupled GAN-network structure, which comprises of two subnets, one for HR and one for LR. The correlation between from the subnet generated features is maximized. Moreover, they considered facial attributes, by implicitly matching facial attributes for both resolutions. 

\subsubsection*{Non-Transformation-based Methods} 
Those techniques try to directly project features from arbitrary image resolutions into the same space. 

In~\cite{zeng2016towards}, Zeng \etal presented a resolution-invariant deep network and trained it directly with unified LR and HR images. However, they used only resolutions in the range of $24$ to $60$ pixels. 

Massoli \etal~\cite{massoli2020cross} proposed a student-teacher network approach. They showed that their approach can be more effective concerning to preprocessing images with super-resolution techniques. 

The authors of~\cite{mudunuri2018genlr} showed that their DCNN architecture can address the problem of CR face recognition. They came up with a two-branch network architecture with several loss functions, which are trained on scale-invariant features for positive and negative sample pairs.

In~\cite{ge2018low}, Ge \etal focused on low computational costs in low-resolution face recognition. Therefore they introduced a new learning approach via selective knowledge distillation. A two-stream technique (large teacher model and a light-weight student model) is employed to transfer selected knowledge from the teacher model to the student model. 
\section{Experimental Setup}
\label{sec:exp_setup}
In this section, we first describe the structure and training process of the baseline network, which we will utilize for our analysis. Then we take a closer look at the five popular datasets we use for testing. After that, we focus on the downsampling method. Finally, we elaborate on our performance measurement method for face verification accuracy. 

\subsection{Baseline Network}
As our baseline network, we choose a network structure comprising a modified ResNet-50~\cite{he2016deep} as proposed in ArcFace~\cite{deng2019arcface}, pretrained on ImageNet~\cite{russakovsky2015imagenet}, and an ArcFace layer for classification. 

The backbone network (ResNet-50) consists of $4$ blocks, which are repeated several times and containing in total $50$ convolutional layers. The image dimension within the layers is decreasing, and the image depth is increasing from $112\,\times\,112\,\times\,3\px$ input to $4\,\times\,4\,\times\,2048\px$ at the end. 

After flattening this $4\,\times4\,\times\,2048\px$ output from the backbone network, dropout is added. A bottleneck layer (512-dimensional fully connected layer), which represents the extracted features and is used for testing, is added following~\cite{wen2016discriminative, liu2017sphereface,wang2018cosface}. Finally, a fully connected layer with the dimension of the number of identities in our training set ($87$k) is added. We apply Additive Angular Margin Loss~\cite{deng2019arcface} together with a cross-entropy classification loss to the network. 

For training, we select the Microsoft MS1M~\cite{guo2016ms} dataset containing about $5.8\mathrm{M}$ images from about $87\mathrm{k}$ identities. We perform random brightness and saturation variations, left-right flipping, and random cropping of images as data augmentation. All training parameters are set according to~\cite{deng2019arcface} except for a smaller batch-size of $128$ due to hardware limitations. The learning rate is set to $0.01$ and is decreased by a factor of $10$ after epoch $9$ and epoch $13$. In total, we train for $16$ epochs with momentum SGD optimizer. The dropout rate and weight decay are set to $0.5$ and $5\cdot 10^{-4}$, respectively. 

\subsection{Testing Datasets}
\label{subsec:datasets}
For our analysis, we select five popular datasets for testing face verification performance: 
\begin{itemize}
	\item The Labeled Faces in the Wild dataset (LFW)~\cite{lfw_dataset} discovers $13233$ images from $5749$ identities. The evaluation protocol contains $6000$ positive (\ie, same identity) image pairs and $6000$ negative (\ie, different identity) image pairs. 
	\item AgeDB~\cite{moschoglou2017agedb} includes $16488$ images of $568$ various people. It addresses a huge variety of age from 1 to 101 years. The evaluation protocol defines $3000$ positive and $3000$ negative image pairs. 
	\item The CFP-FP~\cite{sengupta2016frontal} dataset protocol defines $3500$ positive and $3500$ negative image pairs and compares frontal with profile images. 
	\item The cross age labeled faces in the wild dataset CALFW~\cite{zheng2017cross} is a variant of LFW with focus on the age gap in positive image samples. It contains $3000$ positive and $3000$ negative image pairs. 
	\item The CPLFW~\cite{zheng2018cross} in contrast to CALFW addresses pose variations and selects $3000$ positive and $3000$ negative image pairs with high pose variations within each image pair. 
\end{itemize}

We use the aligned and cropped to $112\,\times\,112\,\times\,3\px$ version of all above-mentioned testing datasets. In this paper, we exclusively deal with images having equal width and height. For simplicity, we denote the image resolution by naming only the first dimension, \ie, a resolution of $112\px$ defines a $112\,\times\,112\,\times\,3\px$ image.

\subsection{Reduction of Image Resolution}
\label{subsec:downsampling}
The baseline network requires HR input images $\im{HR}$ of the size $112\px$. We simulate a resolution-reduction by performing the following two steps. 1) Downsample $\Fdown{r}(\cdot)$ images to an image dimension $r$ in pixels. 2) Subsequently, upsample $\Fup{r}(\cdot)$ those images back to the original image dimension and denote the resulting low-resolution images as $\im{LR}$. The complete process can be formulated as follows:
\begin{equation}
	\im{LR} = \Fup{112}(\Fdown{r}(\im{HR}))
	\label{equ:1}
\end{equation}
 For both sampling processes, bicubic interpolation~\cite{keys1981cubic} is applied. To reduce unwanted artifacts, typically stemming from the downsampling process, standard antialiasing techniques are also involved. In \cref{subsec:res_data}, we further investigate these effects.

\begin{figure}[h]
	\centering
	\includegraphics[width=\textwidth]{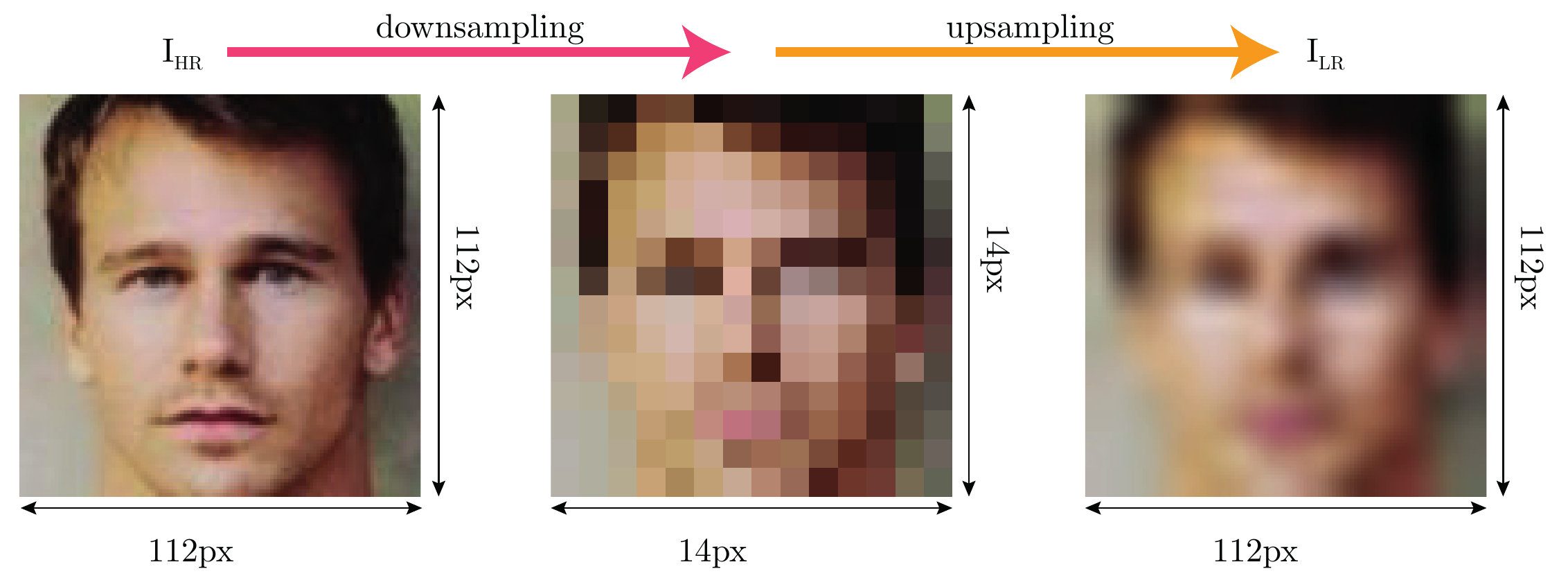}
	\caption{Illustration of the bicubic down- and up-sampling process to reduce the image resolution but keep the image dimension.}
	\label{fig:sampling_overview}
\end{figure}

\cref{fig:sampling_overview} illustrates the synthetic image resolution reduction. The left image $\im{HR}$ is a sample taken from the MS1M dataset with a resolution $r = 112$. In the center one can see the downsampled image $\Fdown{14}(\im{HR})$ with image dimension $r =14$. Finally, the upsampled image $\im{LR}$ is shown on the right and has qualitatively considered an image resolution of $r =14$ but technically the same image dimension as the $\im{HR}$ image. As one can see, all the high-frequency information is removed by this synthetically resolution-reduction. Simultaneously, the image dimension is equal to the original image, which is the required image size for our networks. 

\subsection{Accuracy in Face Verification}
\label{subsec:acc}
We report accuracy in all experiments, which denotes the face recognition rate in terms of face verification. To calculate the accuracy value for a given dataset, we first take the cosine-distances $\dist$ between features of every image pair ($\im{1}, \im{2}$) extracted from a model $\M(\cdot)$ according to $N$ image pairs defined in the specific evaluation protocol for each dataset respectively:

\begin{equation}
	\dist = 1-\sum\biggl(\frac{\M(\im{1})}{\norm{\M(\im{1})}^2} \odot \frac{\M(\im{2})}{\norm{\M(\im{2})}^2}\biggl)
	\label{equ:2}
\end{equation}

Then, we use 10-fold cross-validation to find optimal thresholds that can separate feature distances of positive pairs (\ie, same identity) from negative pairs (not same identity). The number of correctly identified positive samples and negative samples are then named as true positives $TP$ and true negatives $TN$. We then calculate an accuracy score $Acc$ as follows:

\begin{equation}
	Acc = \frac{TP + TN}{N}
	\label{equ:3}
\end{equation}

For all experiments in the CR scenario we generate two evaluation datasets by flipping the pairwise matching resolution from 
$$(\M(\Fup{112}(\Fdown{r}(\im{HR,1}))),\M(\im{HR,2}))$$ to $$(\M(\im{HR,1})\M(\Fup{112}(\Fdown{r}(\im{HR,2})))$$

We then calculate the accuracy score for both test datasets and then take the mean. 

\section{Analysis of Image Resolution Susceptibility}
\label{sec:analysis}
In this section, we first investigate the effect by reducing the resolution across five test datasets. Then, we investigate the performance of the baseline network under LR conditions in CR and ER scenarios. Afterward, we take a closer look at the extracted features, especially at the cosine distance between the image pairs, which is used to classify them as positive (same identity) or negative (different identity). 

\subsection{Effects of Resolution-Reduction on Datasets}
\label{subsec:res_data}
\begin{figure}[t]
	\centering
	\includegraphics[width=\textwidth]{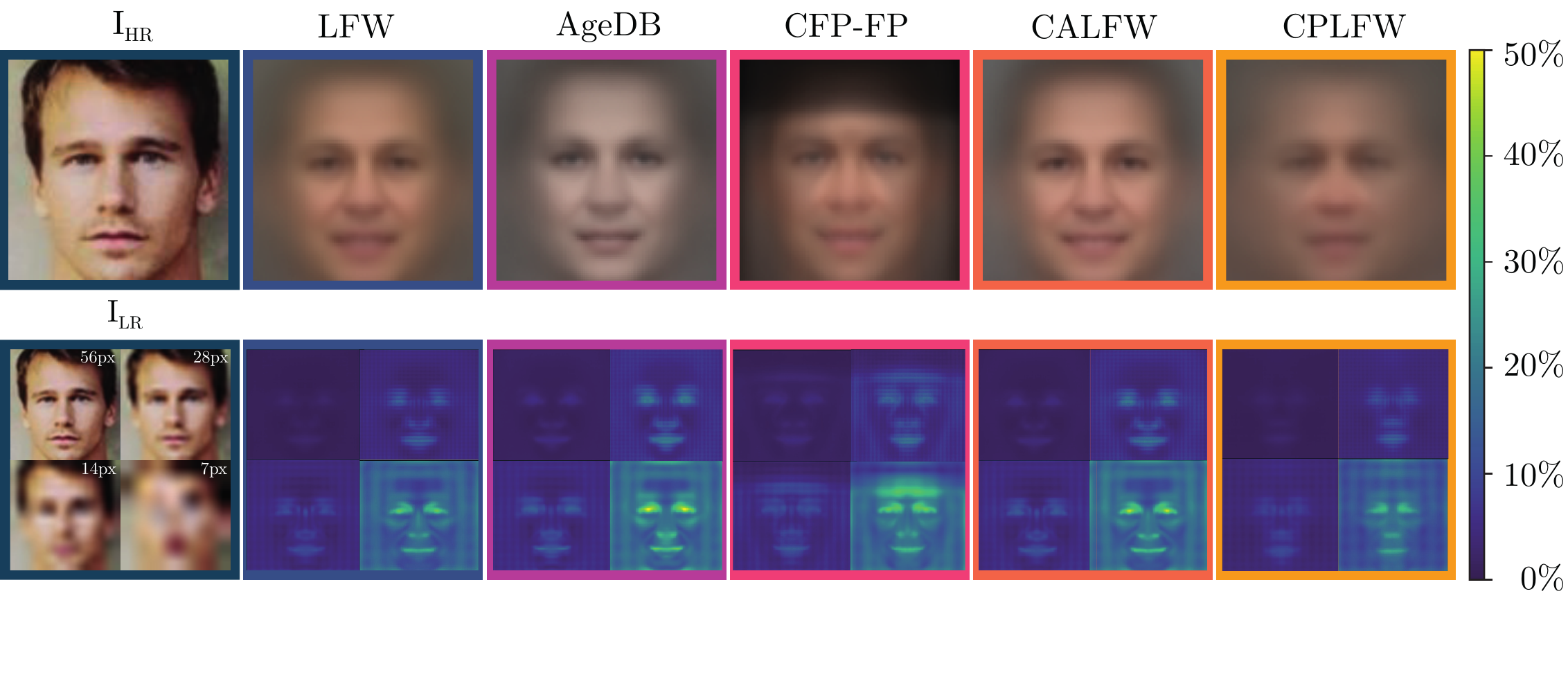}
	\caption{Illustration of the absolute pixel differences after the resolution-reduction process in comparison to the original HR images for several datasets. The left column illustrates the synthetically resolution-reduced images for one sample image picked from the MS1M datasets. The first row of all other images represents the mean image $\im{HR}^{mean}$ for several datasets. Below are the pixel difference images $\Diff{r}^{mean}$ for four specific resolutions $r \in \{7,14,28,56\}$, for the same resolutions as in left-most column.}
	\label{fig:downsample_mean_vis}
\end{figure}

To get a better insight of what is exactly happening when performing the synthetically reduction of image resolution, we elaborate on the difference between testing datasets. Hence, we calculate a mean image across the whole dataset and then highlight the mean pixel difference between LR and HR images. In \cref{fig:downsample_mean_vis} the left column shows a HR sample image $\im{HR}$ from MS1M and its corresponding reduced-resolution images $\Fup{112}(\Fdown{r}(\im{HR}))$ for four resolutions $r \in \{7,14,28,56\}$. All other columns show in the first row the mean HR image $\im{HR}^{mean}$ for several datasets, which is derived as follows:

\begin{equation}
	\im{HR}^{mean} = \frac{\sum_{i=1}^{N}\im{HR,i}}{N}
	\label{equ:4}
\end{equation}

where $N$ denotes the number of elements of the dataset. 

Below each mean HR image, we denote the mean absolute pixel differences $\Diff{r}$ between synthetically reduced images $\im{LR,r}$, and original $\im{HR}$ images across each dataset. We retrieve those images for four resolutions $r \in \{7,14,28,56\}$ according to:

\begin{equation}
	\Diff{r}^{mean} = \frac{\sum_{i=1}^{N}\biggl(\Bigl| \Fup{112}(\Fdown{r}(\im{HR,i})) - \im{HR,i} \Bigl|\biggl)}{N}
	\label{equ:5}
\end{equation}

As expected, eye, nose, and mouth regions are heavily affected by the resolution reduction process in all datasets. High detail information in those regions is lost. The maximum derivation of a single LR image pixel concerning its counterpart pixel in the HR image is about $50\%$. There are slightly visible artifacts in a grid style manner occurring in all pixel-difference images. These might still be some aliasing artifacts, which could not entirely be removed by the antialiasing method of the bicubic interpolation algorithm. The mean dataset HR images are quite different across all datasets. One can clearly see that pose variations in CPLFW dataset result in more blurred areas of the image. In contrast, the CALFW and LFW dataset images seem to be very accurately aligned and show almost a clear and detailed average face. Interestingly, the background in the CFP-FP dataset is very dark compared to other datasets. Also, the pose variation can be seen in the average face. Some ghosting effects are present in that image, too. In terms of pixel derivations, one can see that high-frequency information mainly in the region of eyes, nose, and mouth is lost during the resolution-reduction process. These are valuable information for face recognition. Thus, face verification performance is worse as we later see in the next section. All datasets show the same pattern with respect to their mean absolute pixel difference images. The variation is increasing for lower resolutions. This leads to the conclusion, that the image quality within each dataset is approximately equal. 

\subsection{Face Verification Accuracy}
\begin{figure}[t]
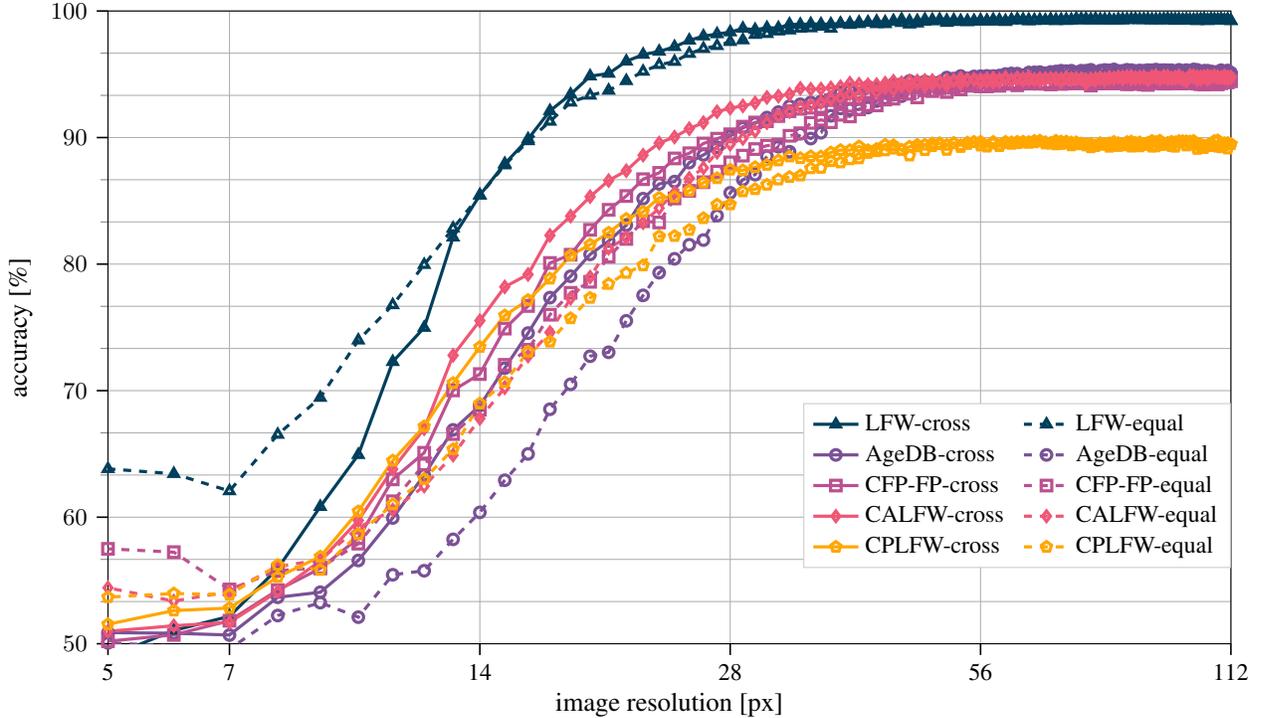

	\centering
	\setlength\figureheight{10cm}
	\setlength\figurewidth{1.0\columnwidth}
	\include{figures/acc_all_lines}
	\caption{Face verification accuracy across several datasets for different image resolutions.}
	\label{fig:acc_datasets}
\end{figure}

\Cref{fig:acc_datasets} visualizes the accuracy for each of the in \cref{subsec:datasets} introduced dataset across different image resolutions in CR (solid lines) and ER (dotted lines) scenario. The performance clearly drops as one would expect for lower resolutions. The best performance overall, but also the largest gap between CR and ER verification accuracy holds for the LFW dataset. In contrast to that, the baseline achieves worse performance on the CPLFW dataset, especially in the ER scenario. A reason for this can be the large pose variations, which are not occurring in the training dataset. 

For a better understanding, what reasons are causing this large decrease of accuracy, we take a closer look at the extracted features from our baseline model in the next section.

\subsection{Feature Distances}
\label{subsec:feat_dual_res}
\begin{figure}[t]
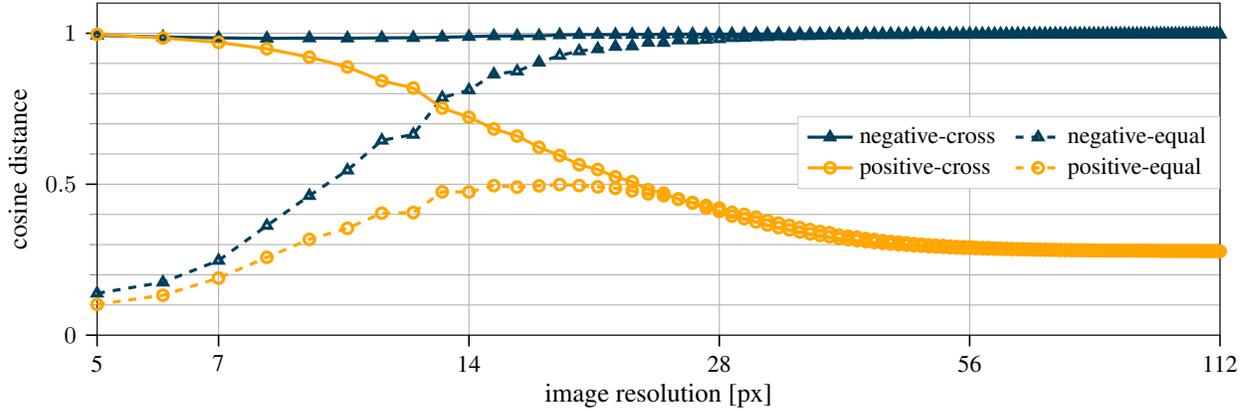

	\centering
	\setlength\figureheight{6cm}
	\setlength\figurewidth{1.0\columnwidth}
	\include{figures/avg_dist_lines}
	\caption{Average cosine feature distances between image pairs for positive (circles) and negative (triangles) \textit{pairs} in the LFW dataset.}
	\label{fig:mean_dist_same_cross}
\end{figure}

The distance between both feature vectors for a test image pair is, according to \cref{equ:2}, crucial for the verification accuracy. Hence, we plot in \cref{fig:mean_dist_same_cross} the average feature distance for all positive image pairs and all negative image pairs in the LFW dataset for a given resolution. As stated in the previous section, we also consider this in the CR and ER scenario. One can divide the behavior roughly into three sections: 1) For high resolutions above about $60\px$, feature distances between positive and negative image pairs seem to be independent of the image resolution. The average distance between positive pairs is quite low about $0.3$ and the distance for negative pairs is about $1.0$, which means that the high dimensional feature vectors are almost orthogonal. 2) Between resolutions of about $60\px$ and $20\px$, which can be considered as mid-range resolutions, in both CR and ER scenarios, the distance of positive image pairs tends to increase. In contrast, the distance for negative image pairs stays at the same level. The most considerable mean feature distance for positive pairs in the ER scenario can be found at image resolutions about $19\px$ with a cosine distance of approximately $0.5$. 3) The last section can be considered as low resolutions less than $20\px$. Interestingly, distances for both scenarios show a contrary behavior. On the one hand, the mean feature distance for CR positive pairs is increasing towards 1. This is consistent with the accuracy decreasing towards about $50\%$, which is in terms of verification, merely guessing. On the other hand, in the ER case, positive and negative feature distances decreasing towards $0.1$. This also coincides with low accuracy scores in that resolution range. 

\begin{figure}[b]
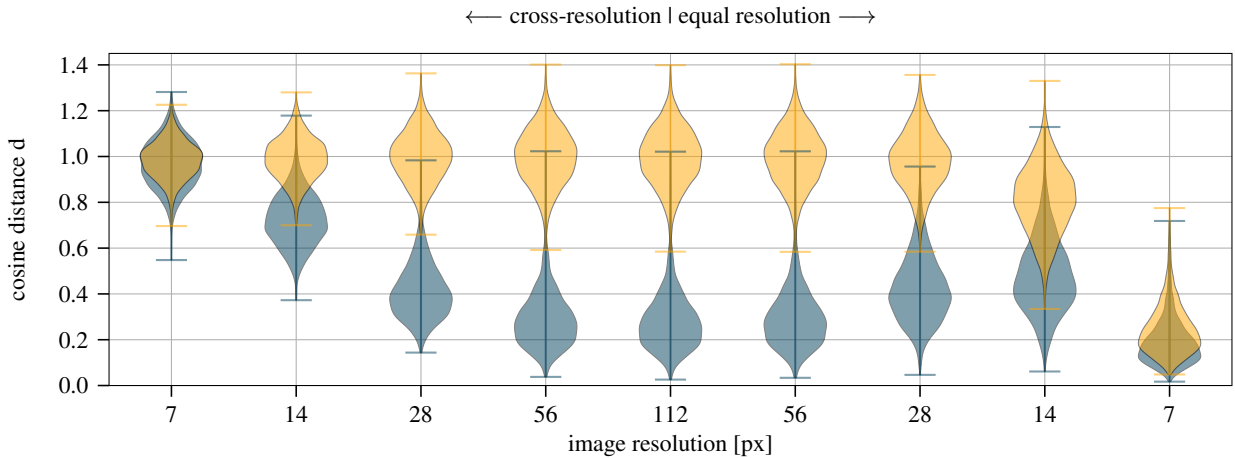

	\centering
	\setlength\figureheight{6cm}
	\setlength\figurewidth{1.0\columnwidth}
	\include{figures/violin_plot_baseline}
	\caption{Cosine feature distance distributions for positive (blue) and negative (yellow) CR (left) and ER (right) pairs in the LFW dataset. Five different resolutions are shown for our baseline model.}
	\label{fig:violin_baseline}
\end{figure}

For the CR scenario, we conduct that our network is not able to extract accurate features for the very low resolution images. Hence, this results in a large distance between features because the HR image features are still very distinctive. However, in the ER scenario both images are somehow unfamiliar to the network and the extracted features are pretty similar. To underline this statement and analyze the distribution of features more fine-grained, we visualize the distributions of positive and negative feature distances for five specific image resolutions ($7\px$, $14\px$, $28\px$, $56\px$, and $112\px$). \Cref{fig:violin_baseline} is capturing those cosine feature distance distributions for the LFW dataset. 

The center violin plots represent the feature distance distribution for HR image pairs. Distances for positive and negative image pairs are clearly distinguishable. The positive distances are mainly in a range between $0.1$ and $0.6$, whereas negative distances are mostly in the field of $0.6$ and $1.4$. Both classes can be separated effectively with a threshold of about $0.6$, and thus, the accuracy for only HR images is best (\cf \cref{fig:acc_datasets}). To the left side, distributions for the CR scenario are shown. On the right side, ER feature distributions are plotted. In both procedures, for a image resolution of $56\px$ no significant difference can be noticed. Interestingly, the peak feature distance for positive image pairs even exceeds the maximum distance for negative pairs in the CR scenario at very low resolution $5\px$. In other words, the resolution has a more enormous impact on the distance than the identity itself. The gap between CR and ER accuracy for very low resolutions is therefore reasonable. Despite of the small distances for both kinds of image pairs in the ER case, still more positive feature distances have a smaller value. This behavior explains a higher accuracy for very low resolutions in the ER scenario compared to CR scenario. Further experiments with CFP-FP, AgeDB, CALFW, and CPLFW datasets show the same trend. 

\section{Proposed Network Structures and Training Methods}
\label{sec:proposed_networks}
To improve the separability between features of positive and negative image pairs, and hence, the accuracy, we pursue two intuitive non-transformation-based methods: 
\begin{itemize}
	\item CR batch training (\BT) with the same network structure as the baseline.
	\item Siamese network CR training (\ST) consisting of- and connecting two baseline architectures.
\end{itemize}
Both approaches are illustrated in \cref{fig:overview_methods}. The following two sections will elaborate on them more in detail. 

\subsection{Cross Resolution Batch Training}
Motivated by~\cite{zeng2016towards,massoli2020cross}, we first propose a straightforward batch CR training approach (\BT) to tackle the image resolutions. The left part of \cref{fig:overview_methods} illustrates this approach. The architecture is equivalent to our baseline network. Instead of applying only HR images, which we did for the baseline training, we now randomly select half of the HR images per batch and synthetically reduce their resolution (\cf \cref{subsec:downsampling}) of those images. Done that, each batch is containing HR and LR images in the same ratio. In the two-resolution training, we train several specific networks specializing each on a particular resolution. We name these models according to the following rule: \BT$-r$ where $r$ denotes the specific LR value during training. To do a more fine-grained analysis at relatively low resolutions we are using the following values for $r\in[5, 22]$ and according to \cref{subsec:feat_dual_res} $r\in\{28, 56\}$. As we only use half of the images per batch for resolution reduction, all networks still see HR images and learn to extract features for HR and LR images at the same gradient updating step. For all CR batch trainings we use the MS1M dataset and train in total for $16$ epochs. All training parameters are set according to our baseline for a fair comparison.

\subsection{Siamese Network Cross Resolution Training}
\begin{figure}[h]
	\centering
	\includegraphics[width=\textwidth]{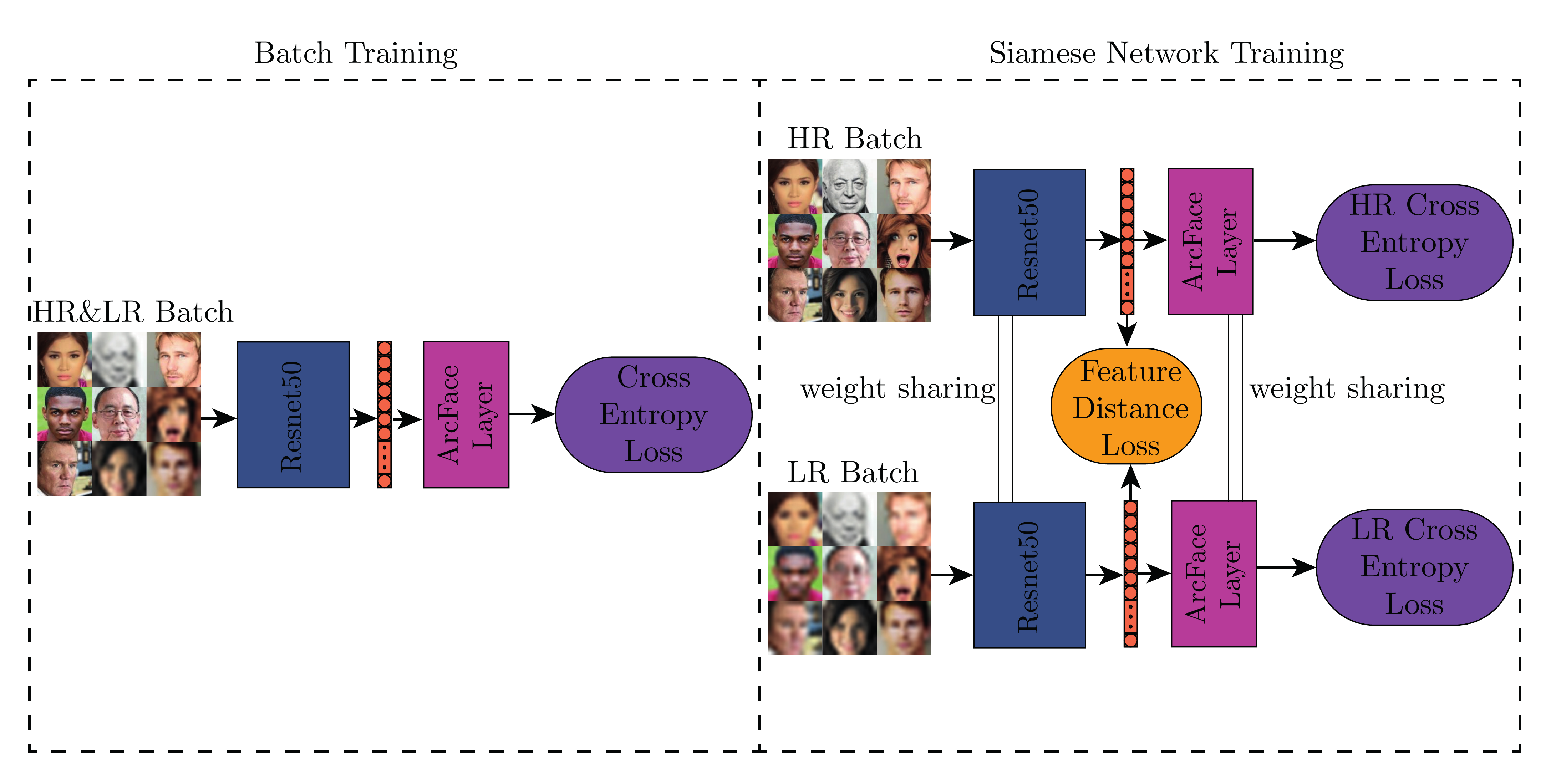}
	\caption{Overview of our proposed methods. The left part shows the CR batch training method (\BT). The right part shows the siamese network (\ST) CR training approach.}
	\label{fig:overview_methods}
\end{figure}

Inspired by Tang \etal~\cite{tang2019finger}, we propose a siamese network CR training (\ST). Therefore, we construct a siamese network structure, as shown in \cref{fig:overview_methods} (right). Generally, each branch of the network is responsible for a specific resolution and all branches share their weights. Thus, the number of parameters is still equal to the baseline. Training with exactly two resolutions HR and LR, requires a two-branch network. Our goal is, that the network projects features from both branches for the same image $\im{}$ in different resolutions, $\im{HR}$ and $\im{LR}$ closely together. To enforce this, we add a new loss function to the network, which penalizes the cosine distance between both features. The cosine distance seems reasonable, since we also evaluate the performance later by calculating cosine distance between image pairs. 
Let $\M(\cdot)$ be the arcface-network then is $\M(\im{HR})$ a feature from the high-resolution branch. The low-resolution branch generates features $\M(\im{LR})$ for a particular resolution respectively. Our feature distance loss $\loss{dist}$ is then:

\begin{equation}
		\loss{dist} = 1-\sum\biggl(\frac{\M(\im{HR})}{\norm{\M(\im{LR})}^2} \odot \frac{\M(\im{HR})}{\norm{\M(\im{LR})}^2}\biggl)
		\label{equ:6}
\end{equation}

For both branches we calculate the cross-entropy classification loss $\loss{ce}^{\mathrm{HR}}$ and  $\loss{ce}^{\mathrm{LR}}$, respectively. We weigh all three losses approximately equally, we add a factor of $25$ to the feature-distance loss. Finally, we conclude the total loss function $\loss{}$ for the siamese training approach as follows: 

\begin{equation}
	\loss{} = \loss{ce}^{\mathrm{HR}} + \loss{ce}^{\mathrm{LR}} + 25 \cdot \loss{dist}
	\label{equ:7}
\end{equation}

Due to the siamese network architecture, the training time is about double in the two resolution training scenario. Both images, HR and LR, need to be inferenced through each branch. Therefore, we select the following resolutions $r \in \{5, 6, 7, 8, 12, 14, 20, 28, 56\}$ to train specific resolution models. For all siamese CR trainings we use MS1M dataset and train for $16$ epochs. All training parameters are set according to our baseline to enable a fair comparison.

\section{Experimental Results}
\label{sec:experimental_results}
In this section, we present and discuss the results of our proposed approaches. Firstly, we focus on the two-resolution scenario, \ie, HR ($112\px$) and one specific LR. Secondly, focus on training with multiple image resolutions simultaneously, \ie, HR ($112\px$) and multiple LRs ($7\px$, $14\px$, $28\px$, $56\px$ and $112\px$) in one training. We analyze the accuracy on five popular datasets and compare the distances of the resulting features for all methods. Moreover, we introduce a new evaluation protocol to measure the performance of a model for multiple resolutions in the test dataset. We are closing this section with a comparison of all in this paper proposed methods, especially concerning the differences in accuracy and training time. 

\subsection{Two Resolution Training Scenario}
According to \cref{sec:proposed_networks}, we now analyze the CR batch training approach \BT and the siamese network CR training approach \ST with respect to the face verification accuracy on five popular datasets. In this two-resolution training scenario, we train each model with exactly two specified resolutions and compare the results to the baseline network concerning accuracy and feature distances.

\subsubsection*{Face Verification Accuracy}

\begin{figure}[t]
	\centering
	\setlength\figureheight{6cm}
	\setlength\figurewidth{1.0\columnwidth}
	\pgfplotsset{every tick label/.append style={font=\footnotesize }}
\begin{tikzpicture}

\definecolor{color1}{HTML}{003f5c}
\definecolor{color2}{HTML}{374c80}
\definecolor{color3}{HTML}{7a5195}
\definecolor{color4}{HTML}{bc5090}
\definecolor{color5}{HTML}{ef5675}
\definecolor{color6}{HTML}{ff764a}
\definecolor{color7}{HTML}{ffa600}

\begin{axis}[
height=\figureheight,
legend cell align={left},
legend entries={{baseline},{\BT},{\ST}},
legend columns=1, 
legend style={
at={(1,0.25)}, 
anchor=east, 
font=\footnotesize ,
/tikz/column 2/.style={column sep=5pt},
draw=white!80.0!black},
tick align=outside,
ylabel near ticks,
tick pos=left,
width=\figurewidth,
x grid style={lightgray!92.02614379084967!black},
xlabel={image resolution [px]},
xmajorgrids,
xmin=5, 
xmax=112,
xminorgrids,
xmode=log,
xtick={5,7,14,28,56,112}, 
xticklabels={5,7,14,28,56,112},
every major tick/.style={black, semithick},
y grid style={lightgray!92.02614379084967!black},
ylabel={accuracy [\%]},
ymajorgrids,
ymin=50, 
ymax=100,
yminorgrids,
minor y tick num=2
]

\addplot [very thick, color1, mark=triangle, mark size=2, mark options={solid}]
table [row sep=\\ ]{
5	50.508	\\
6	51.324	\\
7	51.838	\\
8	54.65	\\
9	56.854	\\
10	60.01	\\
11	64.696	\\
12	67.514	\\
13	72.486	\\
14	74.912	\\
15	77.75	\\
16	79.492	\\
17	82.134	\\
18	83.534	\\
19	85.04	\\
20	86.05	\\
21	87.084	\\
22	88.236	\\
23	89.034	\\
24	89.476	\\
25	90.192	\\
26	90.758	\\
27	91.232	\\
28	91.698	\\
29	92.032	\\
30	92.25	\\
31	92.51	\\
32	92.766	\\
33	93.086	\\
34	93.218	\\
35	93.29	\\
36	93.312	\\
37	93.548	\\
38	93.68	\\
39	93.828	\\
40	93.84	\\
41	93.966	\\
42	94.008	\\
43	94.02	\\
44	94.144	\\
45	94.166	\\
46	94.284	\\
47	94.308	\\
48	94.368	\\
49	94.344	\\
50	94.356	\\
51	94.44	\\
52	94.464	\\
53	94.456	\\
54	94.44	\\
55	94.5	\\
56	94.57	\\
57	94.478	\\
58	94.558	\\
59	94.456	\\
60	94.582	\\
61	94.57	\\
62	94.59	\\
63	94.576	\\
64	94.564	\\
65	94.606	\\
66	94.632	\\
67	94.574	\\
68	94.612	\\
69	94.596	\\
70	94.612	\\
71	94.608	\\
72	94.594	\\
73	94.606	\\
74	94.622	\\
75	94.582	\\
76	94.63	\\
77	94.542	\\
78	94.622	\\
79	94.632	\\
80	94.636	\\
81	94.612	\\
82	94.666	\\
83	94.62	\\
84	94.67	\\
85	94.636	\\
86	94.638	\\
87	94.638	\\
88	94.632	\\
89	94.636	\\
90	94.686	\\
91	94.67	\\
92	94.676	\\
93	94.65	\\
94	94.612	\\
95	94.66	\\
96	94.584	\\
97	94.684	\\
98	94.614	\\
99	94.63	\\
100	94.638	\\
101	94.692	\\
102	94.564	\\
103	94.554	\\
104	94.618	\\
105	94.56	\\
106	94.602	\\
107	94.582	\\
108	94.562	\\
109	94.612	\\
110	94.624	\\
111	94.634	\\
112	94.61	\\
};
\addplot [very thick, color4, mark=square, mark size=2, mark options={solid}]
table [row sep=\\ ]{
5	69.664	\\
6	71.696	\\
7	74.104	\\
8	76.822	\\
9	79.03	\\
10	78.36	\\
11	81.384	\\
12	84.156	\\
13	86.594	\\
14	87.828	\\
15	88.814	\\
16	90.012	\\
17	90.744	\\
18	90.938	\\
19	91.6	\\
20	91.964	\\
21	92.036	\\
22	92.24	\\
28	93.398	\\
56	94.554	\\
};
\addplot [very thick, color7, mark=o, mark size=2, mark options={solid}]
table [row sep=\\ ]{
5	67.856	\\
6	70.126	\\
7	72.768	\\
8	75.142	\\
12	80.99	\\
14	84.212	\\
20	91.192	\\
28	93.378	\\
56	94.692	\\
};
\end{axis}
\end{tikzpicture}
	\caption{Average face verification accuracy across five popular datasets for different resolution. The baseline (triangles), the \BT (squares) and \ST (circles) model are compared.}
	\label{fig:improvs}
\end{figure}
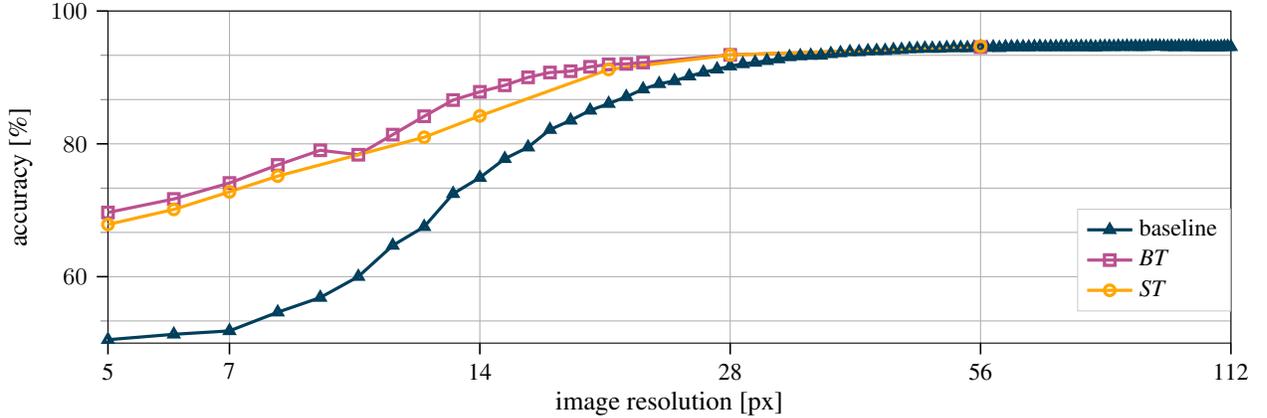

As introduced in \cref{subsec:acc}, accuracy is a common metric to measure the performance of a face verification model. \Cref{fig:improvs} depicts the average face verification accuracy across five common datasets of \BT and \ST model compared to our baseline model. Note that each data point of \BT and \ST represents a different model, which is specifically trained for that resolution. Both approaches clearly outperform the baseline model for low image resolutions. For very low resolutions, \ie, $5\px$ to $8\px$, the performance can be increased from about $50\%$ up to $70\%$. That is equivalent to a relative improvement of $40\%$. Above about $40\px$ image resolution, no significant difference between all approaches is present, which affirms our expectations since the LR images are visually hardly distinguishable from the original images and the absolute pixel difference is very small (\cf \cref{subsec:downsampling}). 

Generally, the performance improvement is increasing with decreasing resolutions. The \BT method performs slightly better than the \ST method, which leads to the conclusion that the siamese approach might concentrate too much on projecting the features of the same image in different resolution to the same space than on classifying the correct identity regardless of the resolution. For applications with a known fixed resolution, a \BT is the better choice. 

Moreover, we compare our results on the very popular LFW dataset with two other approaches (\cf table~\ref{tab:comparison}): The selected knowledge distillation technique proposed by Ge \etal~\cite{ge2018low} and the attribute guided coupled GAN approach introduced by Talreja \etal~\cite{talreja2019attribute}. Our systems clearly outperforms both competitors. However, the comparison to Ge \etal's approach is not be fair, their baseline model (teacher model) only reaches an accuracy of $97.15\%$, which is not comparable to our baseline and state-of-the-art. On the other hand, the model's number of parameters also differs. Both models only trained their models for three different resolutions, which only shows few snapshots and not the whole performance curve. The lowest resolution ($16\px$) is rather high compared to our analysis, so we cannot fully exploit our strengths here. 

\begin{table}[b]
    \centering
    \caption{Face verification accuracy on the LFW dataset. Best performance of each image resolution is marked bold. }
    
\begin{tabular}{cll}
\toprule
\multicolumn{1}{l}{image resolution} & model & accuracy \\
\midrule
\multirow{4}[0]{*}{$64\px$} & \BT-64 (ours) & $\textbf{99.38\%}$ \\
& \ST-64 (ours) & $99.35\%$ \\
& S-64-sc~\cite{ge2018low} & $92.83\%$ \\
& Talreja \etal~\cite{talreja2019attribute} & $94.92\%$ \\
\midrule
\multirow{4}[0]{*}{$32\px$} & \BT-32 (ours) & $\textbf{99.08\%}$ \\
& \ST-32 (ours) & $98.32\%$ \\
& S-32-sc~\cite{ge2018low} & $89.72\%$ \\
& Talreja \etal~\cite{talreja2019attribute} & $91.08\%$ \\
\midrule
\multirow{3}[0]{*}{$16\px$} & \BT-16 (ours) & $\textbf{98.17\%}$ \\
& \ST-16 (ours) & $97.8\%$8 \\
& S-16-sc~\cite{ge2018low} & $85.87\%$ \\
\bottomrule
\end{tabular}
	
    \label{tab:comparison}
\end{table}

\subsubsection*{Feature Distances}
Similar to \cref{sec:analysis}, we pick five different resolutions and take a closer look at the features themselves. To be more precise, we plot the distance distributions for positive and negative image pairs from the LFW dataset. 

\begin{figure}[t]
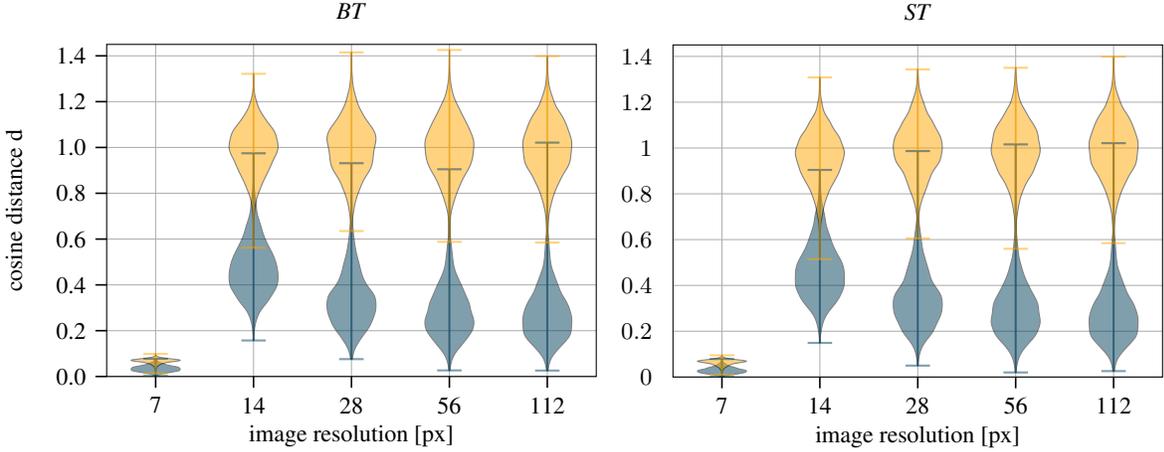

	\centering
	\setlength\figureheight{6cm}
	\setlength\figurewidth{0.49\textwidth}
	\begin{minipage}{0.49\textwidth}
		\include{figures/violin_plot_BT}
	\end{minipage}
	\begin{minipage}{0.49\textwidth}
		\include{figures/violin_plot_ST}
	\end{minipage}
	\caption{Cosine feature distance distributions for positive (blue) and negative (yellow) CR pairs in the LFW dataset. Five different resolutions are compared for \BT (left) and \ST (right).}
	\label{fig:violin_plot_BT_ST}
\end{figure}

In \cref{fig:violin_plot_BT_ST}, on the left side, the distances of extracted features from the \BT model are plotted and on the right side from \ST approach, respectively. Distances of positive and negative pairs are much better separable for low resolutions $14\px$, $28\px$ and $56\px$ than the baseline results (\cf \cref{fig:violin_baseline}). The main difference compared to the baseline is the shift of positive and negative distances to a range of almost $0$ and $0.1$, in the very low-resolution scenario ($7\px$). This is remarkable and shows that both networks are actually learning to project features from very different resolutions into the same space. Although the distances are small, still negative distances are larger than positives and the distributions are separable, which is consistent with the accuracy improvement discussed in the previous section (\cf \cref{fig:improvs}). Furthermore, there is no significant difference between both proposed methods, which is also consistent with the previous section's accuracy values.

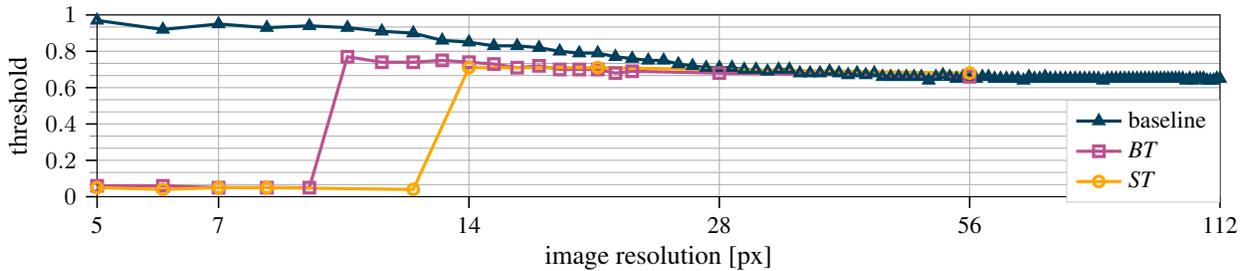
\begin{figure}[b]
	\centering
	\setlength\figureheight{4cm}
	\setlength\figurewidth{1.0\columnwidth}
	\pgfplotsset{every tick label/.append style={font=\footnotesize }}
\begin{tikzpicture}

\definecolor{color1}{HTML}{003f5c}
\definecolor{color2}{HTML}{374c80}
\definecolor{color3}{HTML}{7a5195}
\definecolor{color4}{HTML}{bc5090}
\definecolor{color5}{HTML}{ef5675}
\definecolor{color6}{HTML}{ff764a}
\definecolor{color7}{HTML}{ffa600}

\begin{axis}[
height=\figureheight,
legend cell align={left},
legend entries={{baseline},{\BT},{\ST}},
legend columns=1, 
legend style={
at={(1,0.25)}, 
anchor=east, 
font=\footnotesize ,
/tikz/column 2/.style={column sep=5pt},
draw=white!80.0!black},
tick align=outside,
ylabel near ticks,
tick pos=left,
width=\figurewidth,
x grid style={lightgray!92.02614379084967!black},
xlabel={image resolution [px]},
xmajorgrids,
xmin=5, 
xmax=112,
xminorgrids,
xmode=log,
xtick={5,7,14,28,56,112}, 
xticklabels={5,7,14,28,56,112},
every major tick/.style={black, semithick},
y grid style={lightgray!92.02614379084967!black},
ylabel={threshold},
ymajorgrids,
ymin=0, 
ymax=1,
yminorgrids,
minor y tick num=2
]

\addplot [very thick, color1, mark=triangle, mark size=2, mark options={solid}]
table [row sep=\\ ]{
5	0.97	\\
6	0.92	\\
7	0.95	\\
8	0.93	\\
9	0.94	\\
10	0.93	\\
11	0.91	\\
12	0.9	\\
13	0.86	\\
14	0.85	\\
15	0.83	\\
16	0.83	\\
17	0.82	\\
18	0.8	\\
19	0.79	\\
20	0.79	\\
21	0.77	\\
22	0.76	\\
23	0.75	\\
24	0.75	\\
25	0.73	\\
26	0.72	\\
27	0.71	\\
28	0.71	\\
29	0.71	\\
30	0.7	\\
31	0.7	\\
32	0.69	\\
33	0.7	\\
34	0.7	\\
35	0.68	\\
36	0.68	\\
37	0.68	\\
38	0.69	\\
39	0.68	\\
40	0.67	\\
41	0.68	\\
42	0.67	\\
43	0.68	\\
44	0.66	\\
45	0.66	\\
46	0.66	\\
47	0.66	\\
48	0.66	\\
49	0.66	\\
50	0.64	\\
51	0.66	\\
52	0.67	\\
53	0.66	\\
54	0.65	\\
55	0.66	\\
56	0.65	\\
57	0.65	\\
58	0.66	\\
59	0.66	\\
60	0.65	\\
61	0.65	\\
62	0.65	\\
63	0.65	\\
64	0.65	\\
65	0.64	\\
66	0.66	\\
67	0.65	\\
68	0.65	\\
69	0.66	\\
70	0.65	\\
71	0.65	\\
72	0.65	\\
73	0.65	\\
74	0.65	\\
75	0.65	\\
76	0.65	\\
77	0.65	\\
78	0.65	\\
79	0.65	\\
80	0.65	\\
81	0.64	\\
82	0.65	\\
83	0.65	\\
84	0.65	\\
85	0.65	\\
86	0.65	\\
87	0.65	\\
88	0.65	\\
89	0.65	\\
90	0.65	\\
91	0.65	\\
92	0.65	\\
93	0.65	\\
94	0.65	\\
95	0.65	\\
96	0.65	\\
97	0.65	\\
98	0.65	\\
99	0.65	\\
100	0.65	\\
101	0.65	\\
102	0.64	\\
103	0.65	\\
104	0.64	\\
105	0.65	\\
106	0.65	\\
107	0.65	\\
108	0.64	\\
109	0.64	\\
110	0.64	\\
111	0.65	\\
112	0.65	\\
};
\addplot [very thick, color4, mark=square, mark size=2, mark options={solid}]
table [row sep=\\ ]{
5	0.06	\\
6	0.06	\\
7	0.05	\\
8	0.05	\\
9	0.05	\\
10	0.77	\\
11	0.74	\\
12	0.74	\\
13	0.75	\\
14	0.74	\\
15	0.73	\\
16	0.71	\\
17	0.72	\\
18	0.7	\\
19	0.7	\\
20	0.7	\\
21	0.68	\\
22	0.69	\\
28	0.68	\\
56	0.66	\\
};
\addplot [very thick, color7, mark=o, mark size=2, mark options={solid}]
table [row sep=\\ ]{
5	0.05	\\
6	0.04	\\
7	0.05	\\
8	0.05	\\
12	0.04	\\
14	0.71	\\
20	0.71	\\
56	0.68	\\
};
\end{axis}
\end{tikzpicture}
	\caption{Best threshold values for calculating the CR accuracy on the LFW datasets using baseline (diamonds), \BT (squares) and \ST (circles) models. }
	\label{fig:thresh_analyse}
\end{figure}

To understand and determine the exact resolution at where the feature distances drop so much, we analyze the optimal threshold, which is calculated for the accuracy values (\cf \cref{subsec:acc}). \Cref{fig:thresh_analyse} depicts the threshold values for the baseline, \BT, and \ST model on all tested image resolution in the CR scenario. Thresholds for our baseline model are increasing for lower resolutions. This is consistent with our results in \cref{subsec:feat_dual_res}, where both positive and negative feature distances increase for lower resolutions. Both of our two-resolution training networks show a significant drop, at a resolution of $9\px$ for \BT and approximately $12\px$ for \ST. From these points on, both models are training differently and projecting features for large resolution differences more closely together. There is no connection between this drop in feature distances and the performance of both models conductible. 

\subsection{Multi-Resolution Training Scenario}
\label{subsec:multi_res_nets}
To simulate a more applicable model, which is capable of handling arbitrary resolutions, we propose a multiple-resolution training for both of our previously presented approaches. We train the \BT model with more than two resolutions at the same time by simply randomly pick a different resolution from the numbers $\{7,14,28,56,112\}$ for generating a LR image. Each batch contains multiple LR images with different resolutions and HR images. We find that those five resolutions equally represent the range of image resolutions. This reflects, for example, equivalent distances from subjects to camera in real life. The probability of each resolution is set to be equal. We name this approach \BT-M in the following. 

In the \ST approach, we apply two different methods for training with multiple resolutions simultaneously. First, for the LR branch, we simply randomly pick a resolution from the numbers $\{7,14,28,56\}$ and feed the LR branch with different resolution LR images within each batch. The HR branch always takes $112\px$ images. Hence, we only double the training time but train with in total five different resolutions simultaneously. That approach will be referred to as \ST-M2 in the following. The second method \ST-M2 extends the siamese network to in total to five branches, each branch representing a certain defined resolution ($7\px$, $14\px$, $28\px$, $56\px$, and $112\px$). Consequently, five feature distance losses are calculated each between the HR and the corresponding LR branch. Moreover, we also calculate the cross-entropy loss for each branch. All feature distance losses are weighted with a factor of $3$ to be in the same order of magnitude as the cross-entropy losses. The training time for this experiment is about five times longer than the baseline because it is scaling with the number of defined resolutions for training. 

\subsubsection*{Face Verification Accuracy}
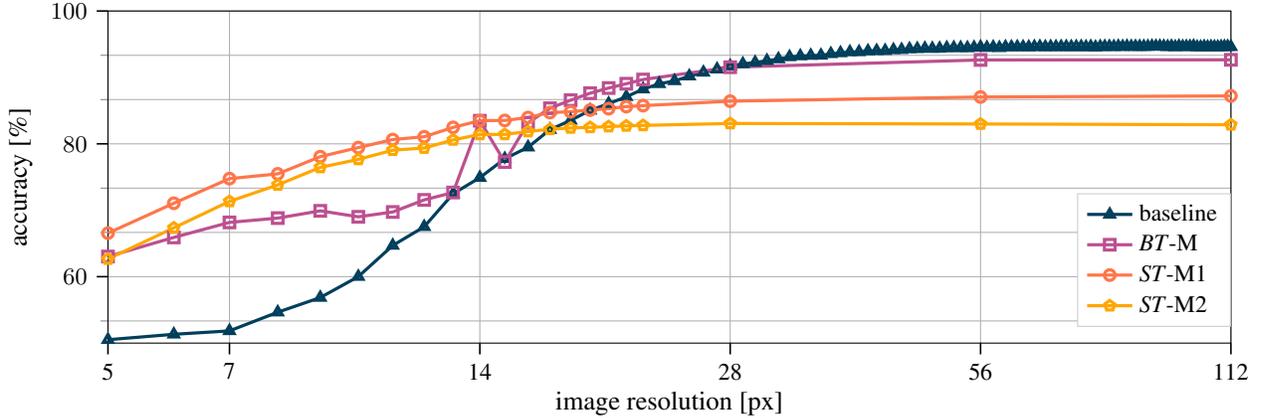
\begin{figure}[t]
	\centering
	\setlength\figureheight{6cm}
	\setlength\figurewidth{1.0\columnwidth}
	\pgfplotsset{every tick label/.append style={font=\footnotesize }}
\begin{tikzpicture}

\definecolor{color1}{HTML}{003f5c}
\definecolor{color2}{HTML}{374c80}
\definecolor{color3}{HTML}{7a5195}
\definecolor{color4}{HTML}{bc5090}
\definecolor{color5}{HTML}{ef5675}
\definecolor{color6}{HTML}{ff764a}
\definecolor{color7}{HTML}{ffa600}

\begin{axis}[
height=\figureheight,
legend cell align={left},
legend entries={{baseline},{\BT-M},{\ST-M1},{\ST-M2}},
legend columns=1, 
legend style={
at={(1,0.25)}, 
anchor=east, 
font=\footnotesize ,
/tikz/column 2/.style={column sep=5pt},
draw=white!80.0!black},
tick align=outside,
ylabel near ticks,
tick pos=left,
width=\figurewidth,
x grid style={lightgray!92.02614379084967!black},
xlabel={image resolution [px]},
xmajorgrids,
xmin=5, 
xmax=112,
xminorgrids,
xmode=log,
xtick={5,7,14,28,56,112}, 
xticklabels={5,7,14,28,56,112},
every major tick/.style={black, semithick},
y grid style={lightgray!92.02614379084967!black},
ylabel={accuracy [\%]},
ymajorgrids,
ymin=50, 
ymax=100,
yminorgrids,
minor y tick num=2
]

\addplot [very thick, color1, mark=triangle, mark size=2, mark options={solid}]
table [row sep=\\ ]{
5	50.508	\\
6	51.324	\\
7	51.838	\\
8	54.65	\\
9	56.854	\\
10	60.01	\\
11	64.696	\\
12	67.514	\\
13	72.486	\\
14	74.912	\\
15	77.75	\\
16	79.492	\\
17	82.134	\\
18	83.534	\\
19	85.04	\\
20	86.05	\\
21	87.084	\\
22	88.236	\\
23	89.034	\\
24	89.476	\\
25	90.192	\\
26	90.758	\\
27	91.232	\\
28	91.698	\\
29	92.032	\\
30	92.25	\\
31	92.51	\\
32	92.766	\\
33	93.086	\\
34	93.218	\\
35	93.29	\\
36	93.312	\\
37	93.548	\\
38	93.68	\\
39	93.828	\\
40	93.84	\\
41	93.966	\\
42	94.008	\\
43	94.02	\\
44	94.144	\\
45	94.166	\\
46	94.284	\\
47	94.308	\\
48	94.368	\\
49	94.344	\\
50	94.356	\\
51	94.44	\\
52	94.464	\\
53	94.456	\\
54	94.44	\\
55	94.5	\\
56	94.57	\\
57	94.478	\\
58	94.558	\\
59	94.456	\\
60	94.582	\\
61	94.57	\\
62	94.59	\\
63	94.576	\\
64	94.564	\\
65	94.606	\\
66	94.632	\\
67	94.574	\\
68	94.612	\\
69	94.596	\\
70	94.612	\\
71	94.608	\\
72	94.594	\\
73	94.606	\\
74	94.622	\\
75	94.582	\\
76	94.63	\\
77	94.542	\\
78	94.622	\\
79	94.632	\\
80	94.636	\\
81	94.612	\\
82	94.666	\\
83	94.62	\\
84	94.67	\\
85	94.636	\\
86	94.638	\\
87	94.638	\\
88	94.632	\\
89	94.636	\\
90	94.686	\\
91	94.67	\\
92	94.676	\\
93	94.65	\\
94	94.612	\\
95	94.66	\\
96	94.584	\\
97	94.684	\\
98	94.614	\\
99	94.63	\\
100	94.638	\\
101	94.692	\\
102	94.564	\\
103	94.554	\\
104	94.618	\\
105	94.56	\\
106	94.602	\\
107	94.582	\\
108	94.562	\\
109	94.612	\\
110	94.624	\\
111	94.634	\\
112	94.61	\\
};
\addplot [very thick, color4, mark=square, mark size=2, mark options={solid}]
table [row sep=\\ ]{
5	63.034	\\
6	65.908	\\
7	68.172	\\
8	68.82	\\
9	69.912	\\
10	69.014	\\
11	69.74	\\
12	71.56	\\
13	72.672	\\
14	83.478	\\
15	77.264	\\
16	83.176	\\
17	85.364	\\
18	86.572	\\
19	87.624	\\
20	88.404	\\
21	89.052	\\
22	89.684	\\
28	91.536	\\
56	92.632	\\
112	92.652	\\
};
\addplot [very thick, color6, mark=o, mark size=2, mark options={solid}]
table [row sep=\\ ]{
5	66.554	\\
6	71.052	\\
7	74.762	\\
8	75.478	\\
9	78.07	\\
10	79.436	\\
11	80.646	\\
12	81.068	\\
13	82.478	\\
14	83.494	\\
15	83.52	\\
16	83.964	\\
17	84.706	\\
18	84.834	\\
19	85.09	\\
20	85.296	\\
21	85.608	\\
22	85.752	\\
28	86.422	\\
56	87.05	\\
112	87.214	\\
};
\addplot [very thick, color7, mark=pentagon, mark size=2, mark options={solid}]
table [row sep=\\ ]{
5	62.658	\\
6	67.35	\\
7	71.334	\\
8	73.816	\\
9	76.45	\\
10	77.66	\\
11	79.044	\\
12	79.37	\\
13	80.594	\\
14	81.416	\\
15	81.42	\\
16	81.856	\\
17	82.188	\\
18	82.394	\\
19	82.476	\\
20	82.602	\\
21	82.702	\\
22	82.776	\\
28	83.066	\\
56	82.99	\\
112	82.872	\\
};
\end{axis}
\end{tikzpicture}
	\caption{Average face verification accuracy across five popular datasets for different image resolutions. The baseline (diamonds), the \BT-M (squares), \ST-M1 (circles), and \ST-M2 (pentagons) model are compared.}
	\label{fig:multiIMP}
\end{figure}

In \cref{fig:multiIMP}, we present the face verification accuracy for \BT-M, \ST-M1, and \ST-M2 model across arbitrary image resolutions. All three approaches perform significantly better than the baseline model in resolutions below $13\px$ and worse above a resolution of $28\px$. Note that there is a significant peak at a resolution of $14\px$, especially for \BT-M. One reason for this could be, that this specific resolution was in the training resolution included. This effect is not visible at a resolution of $7\px$, $28\px$, and $56\px$. Another finding we can conduct is that the siamese network CR training outperforms the CR batch training for low resolutions (approx. below $16\px$) and vice-versa for mid and high resolutions (approx. above $16\px$). At a low resolution of $7\px$ the \ST-M1 model achieves an accuracy score of about $75\%$, that is almost $25\%$ above the baseline performance. At the same time, that approach looses about $8\%$ performance at high resolutions like $56\px$. For a more scale comprehensive performance score, we will introduce three new evaluation protocols in \cref{subsec:eval_proto}. 

\begin{figure}[t]
	\centering
	\setlength\figureheight{8cm}
	\setlength\figurewidth{0.49\columnwidth}
	\begin{minipage}{0.49\columnwidth}
		\pgfplotsset{every tick label/.append style={font=\footnotesize }}
\begin{tikzpicture}

\definecolor{color1}{HTML}{003f5c}
\definecolor{color2}{HTML}{374c80}
\definecolor{color3}{HTML}{7a5195}
\definecolor{color4}{HTML}{bc5090}
\definecolor{color5}{HTML}{ef5675}
\definecolor{color6}{HTML}{ff764a}
\definecolor{color7}{HTML}{ffa600}
\definecolor{color0}{HTML}{000000}

\begin{axis}[
height=\figureheight,
legend cell align={left},
legend entries={{\BT-5-9},{\BT-7},{\BT-M},{baseline},{\ST-M1},{\ST-M2},{\ST-5-8},{\ST-7}},
legend columns=3, 
legend style={
at={(1,1.15)}, 
anchor=east, 
font=\footnotesize ,
/tikz/column 2/.style={column sep=5pt},
draw=white!80.0!black},
tick align=outside,
ylabel near ticks,
tick pos=left,
width=\figurewidth,
x grid style={lightgray!92.02614379084967!black},
xlabel={image resolution [px]},
xmajorgrids,
xmin=5, 
xmax=9,
xminorgrids,
xtick={5,6,7,8,9}, 
xticklabels={5,6,7,8,9},
every major tick/.style={black, semithick},
y grid style={lightgray!92.02614379084967!black},
ylabel={accuracy [\%]},
ymajorgrids,
ymin=48, 
ymax=93,
yminorgrids,
minor y tick num=2
]

\addplot [very thick, color1, mark=cross, mark size=2, mark options={solid}]
table [row sep=\\ ]{
5	82.55	\\
6	84.76	\\
7	87.36	\\
8	89.88	\\
9	91.72	\\
};
\addplot [very thick, color2, mark=square, mark size=2, mark options={solid}]
table [row sep=\\ ]{
5	75.67	\\
6	82.38	\\
7	87.36	\\
8	86.93	\\
9	89.35	\\
};
\addplot [very thick, color3, mark=triangle, mark size=2, mark options={solid}]
table [row sep=\\ ]{
5	71.53	\\
6	75.15	\\
7	77.36	\\
8	77.33	\\
9	77.71	\\
};
\addplot [very thick, color0, mark=*, mark size=2, mark options={solid}]
table [row sep=\\ ]{
5	48.94	\\
6	51.05	\\
7	52.17	\\
8	55.94	\\
9	60.83	\\
};
\addplot [very thick, color4, mark=square, mark size=2, mark options={solid}]
table [row sep=\\ ]{
5	76.78	\\
6	83.23	\\
7	87.64	\\
8	88.18	\\
9	91.28	\\
};
\addplot [very thick, color5, mark=o, mark size=2, mark options={solid}]
table [row sep=\\ ]{
5	71.84	\\
6	79.28	\\
7	85.39	\\
8	87.8	\\
9	90.88	\\
};
\addplot [very thick, color6, mark=pentagon, mark size=2, mark options={solid}]
table [row sep=\\ ]{
5	82.57	\\
6	84.81	\\
7	88.47	\\
8	90.42	\\
};
\addplot [very thick, color7, mark=diamond, mark size=2, mark options={solid}]
table [row sep=\\ ]{
5	76.58	\\
6	82.25	\\
7	88.47	\\
8	87.65	\\
9	88.72	\\
};

\end{axis}
\end{tikzpicture}
	\end{minipage}
	\begin{minipage}{0.49\columnwidth}
		\pgfplotsset{every tick label/.append style={font=\footnotesize }}
\begin{tikzpicture}

\definecolor{color1}{HTML}{003f5c}
\definecolor{color2}{HTML}{374c80}
\definecolor{color3}{HTML}{7a5195}
\definecolor{color4}{HTML}{bc5090}
\definecolor{color5}{HTML}{ef5675}
\definecolor{color6}{HTML}{ff764a}
\definecolor{color7}{HTML}{ffa600}
\definecolor{color0}{HTML}{000000}

\begin{axis}[
height=\figureheight,
legend cell align={left},
legend entries={{\BT-10-18},{\BT-14},{\BT-M},{baseline},{\ST-M1},{\ST-M2},{\ST-12,14},{\ST-14}},
legend columns=3, 
legend style={
at={(1,1.15)}, 
anchor=east, 
font=\footnotesize ,
/tikz/column 2/.style={column sep=5pt},
draw=white!80.0!black},
tick align=outside,
ylabel near ticks,
tick pos=left,
width=\figurewidth,
x grid style={lightgray!92.02614379084967!black},
xlabel={image resolution [px]},
xmajorgrids,
xmin=10, 
xmax=18,
xminorgrids,
xtick={10,11,12,13,14,15,16,17,18}, 
xticklabels={10,11,12,13,14,15,16,17,18},
every major tick/.style={black, semithick},
y grid style={lightgray!92.02614379084967!black},
ylabel={accuracy [\%]},
ymajorgrids,
ymin=63, 
ymax=100,
yminorgrids,
minor y tick num=2
]

\addplot [very thick, color1, mark=cross, mark size=2, mark options={solid}]
table [row sep=\\ ]{
10	93.13	\\
11	95.13	\\
12	96.64	\\
13	97.5	\\
14	97.72	\\
15	97.98	\\
16	98.17	\\
17	98.38	\\
18	98.59	\\
};
\addplot [very thick, color2, mark=square, mark size=2, mark options={solid}]
table [row sep=\\ ]{
10	90.04	\\
11	93.23	\\
12	94.19	\\
13	95.65	\\
14	97.72	\\
15	97.3	\\
16	97.97	\\
17	98.07	\\
18	98.43	\\
};
\addplot [very thick, color3, mark=triangle, mark size=2, mark options={solid}]
table [row sep=\\ ]{
10	73.42	\\
11	74.11	\\
12	78.33	\\
13	80.19	\\
14	97.34	\\
15	90.31	\\
16	96.9	\\
17	97.8	\\
18	97.92	\\
};
\addplot [very thick, color0, mark=*, mark size=2, mark options={solid}]
table [row sep=\\ ]{
10	64.93	\\
11	72.28	\\
12	74.98	\\
13	82.12	\\
14	85.45	\\
15	87.93	\\
16	89.88	\\
17	92.11	\\
18	93.41	\\
};
\addplot [very thick, color4, mark=square, mark size=2, mark options={solid}]
table [row sep=\\ ]{
10	92.1	\\
11	93.44	\\
12	93.31	\\
13	95	\\
14	95.6	\\
15	95.7	\\
16	96.13	\\
17	96.29	\\
18	96.51	\\
};
\addplot [very thick, color5, mark=o, mark size=2, mark options={solid}]
table [row sep=\\ ]{
10	91.45	\\
11	93.12	\\
12	93.63	\\
13	94.74	\\
14	95.15	\\
15	95.18	\\
16	95.23	\\
17	95.31	\\
18	95.53	\\
};
\addplot [very thick, color6, mark=pentagon, mark size=2, mark options={solid}]
table [row sep=\\ ]{
12	94.12	\\
14	96.55	\\
};
\addplot [very thick, color7, mark=diamond, mark size=2, mark options={solid}]
table [row sep=\\ ]{
10	83.01	\\
11	89.48	\\
12	91.73	\\
13	94.6	\\
14	96.55	\\
15	96.48	\\
16	97.01	\\
17	97.37	\\
18	97.47	\\
};

\end{axis}
\end{tikzpicture}
	\end{minipage}
	\caption{Accuracy for LFW dataset in the region of $5\px$ and $9\px$ (left) and $10\px$ and $18\px$ (right.)}
	\label{fig:lines_ds_sizes_diff}
\end{figure}
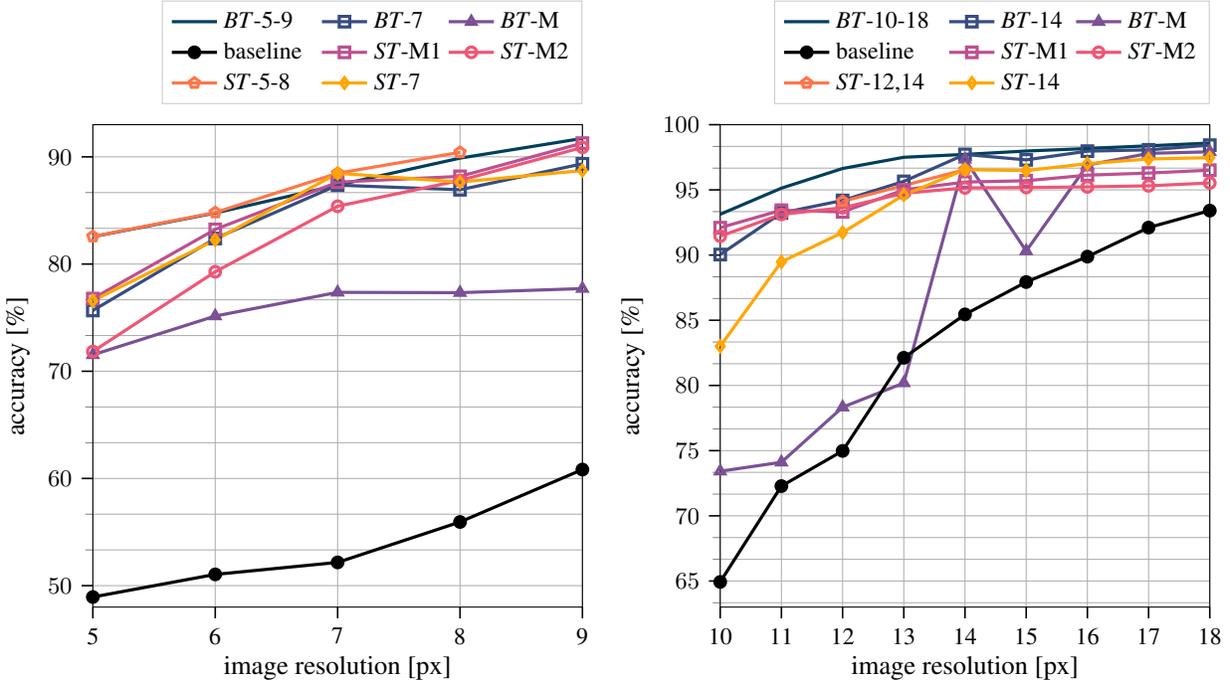

In \cref{fig:lines_ds_sizes_diff} we investigate the performance at and close to two selected resolutions $7\px$ and $14\px$. On the left side we can see that \BT-7 and \ST-7 optimized the performance exactly for the $7\px$ resolution, hence they perform worse in the neighboring regions. Obviously, \BT-5-9 and \ST-5-8, which both represent specific resolution trained models, performing best at each scale. This is reasonable due to the training with that particular image resolution. The performance loss for all multiple-resolution trained approaches (\BT-M, \ST-M1, and \ST-M2) is compensated by the benefit of having a single model for arbitrary resolutions. 
The right part of \cref{fig:lines_ds_sizes_diff} shows an excerpt of resolutions from $10\px$ and $18\px$. In other words, it illustrates the neighboring region of a $14\px$ image resolution. Here, the wave effect of \BT-14 and \ST-14 is also present, meaning that those two models perform relatively best on exactly $14\px$ resolution. Remarkable is a significant peak in accuracy for \BT-M model at the focused resolution of $14\px$. The for each specific trained two-resolution networks \BT-10-18 is performing best on all scales. 

\subsubsection*{Feature Distances}
\begin{figure}[b]
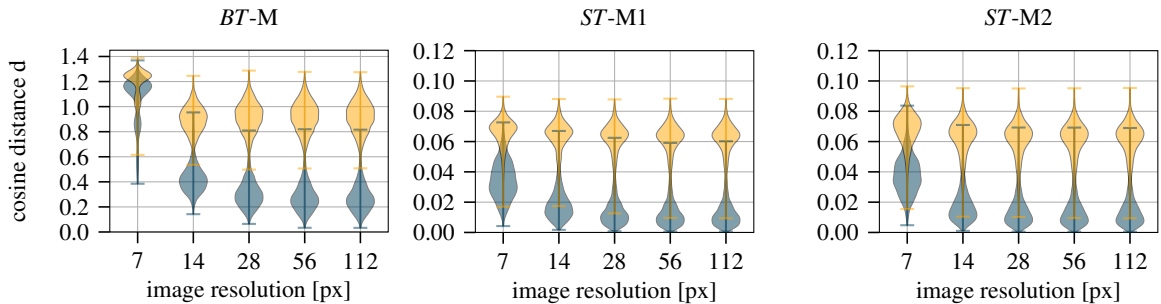

	\centering
	\setlength\figureheight{4cm}
	\setlength\figurewidth{0.32\textwidth}
	\begin{minipage}{0.32\textwidth}
		\include{figures/violin_plot_BT-M}
	\end{minipage}
	\begin{minipage}{0.32\textwidth}
		\include{figures/violin_plot_ST-M1}
	\end{minipage}
	\begin{minipage}{0.32\textwidth}
		\include{figures/violin_plot_ST-M2}
	\end{minipage}
	\caption{Cosine feature distance distributions for positive (blue) and negative (yellow) CR pairs in the LFW dataset. Five different resolutions are compared for \BT-M (left), \ST-M1 (center), and \ST-M2 (right). }
	\label{fig:violin_multi}
\end{figure}

Interestingly, in terms of feature distance distributions (\cf left part of \cref{fig:violin_multi}), the multi-resolution batch training is not behaving similarly to the two resolution batch training. Specifically for \BT, at very low resolutions ($7\px$), the feature distance distributions for positive and negative pairs are even larger than for all other resolutions. This fits together with the \BT accuracy at that scale (\cf \cref{fig:lines_ds_sizes_diff}). In contrast to the two-resolution case, both siamese training approaches (\ST-M1 and \ST-M2) project features for all resolutions more closely together. We conduct this from very low distances across all scales (middle and right part in \cref{fig:violin_multi}). The maximum feature distance for both approaches is about $0.1$. 

\subsection{Evaluation Protocols for Multiple Resolutions}
\label{subsec:eval_proto}
Evaluation protocols for common public datasets are not taking the image resolution into account. In the previous sections, we only considered a specific image resolution to calculate the face verification accuracy. With single networks (\cf \cref{subsec:multi_res_nets}) capable of handling arbitrary image resolutions at once, there is a need for a more meaningful evaluation considering multiple resolutions. Therefore, we propose evaluation protocols for all five datasets. We focus on three different resolution ranges:
\begin{itemize}
	\item Low resolution: Resolutions in the range of $5\px$ to $10\px$ are randomly picked with each with the same probability for the LR images in the CR evaluation. That means we derive a list containing the resolution for each image in each pair for the corresponding dataset.
	\item Mid resolution:  Resolutions in the range of $10\px$ to $40\px$ are used for selection.
	\item High resolution:  Resolutions in the range of $40\px$ to $112\px$ are used for selection.
\end{itemize}

Moreover, we generate a protocol that takes every single resolution in the range of $5\px$ to $112\px$ into account. This will be referred to as \textit{all\_res} in table~\ref{table:1}. All evaluation protocols are published and available on GitHub\footnote{\href{https://github.com/Martlgap/btm-stm}{https://github.com/Martlgap/btm-stm}}. 

\subsection{Comparison of the proposed Methods}
\begin{figure}[b]
	\centering
	\setlength\figureheight{6cm}
	\setlength\figurewidth{1.0\columnwidth}
	\pgfplotsset{every tick label/.append style={font=\footnotesize }}
\begin{tikzpicture}

\definecolor{color1}{HTML}{003f5c}
\definecolor{color2}{HTML}{374c80}
\definecolor{color3}{HTML}{7a5195}
\definecolor{color4}{HTML}{bc5090}
\definecolor{color5}{HTML}{ef5675}
\definecolor{color6}{HTML}{ff764a}
\definecolor{color7}{HTML}{ffa600}

\begin{axis}[
height=\figureheight,
legend cell align={left},
legend entries={{\BT-56},{baseline},{\BT-7},{\ST-7},{\ST-56}},
legend columns=2, 
legend style={
at={(1,0.25)}, 
anchor=east, 
font=\footnotesize ,
/tikz/column 2/.style={column sep=5pt},
draw=white!80.0!black},
tick align=outside,
ylabel near ticks,
tick pos=left,
width=\figurewidth,
x grid style={lightgray!92.02614379084967!black},
xlabel={epoch},
xmajorgrids,
xmin=1, 
xmax=16,
xminorgrids,
xtick={1,2,3,4,5,6,7,8,9,10,11,12,13,14,15,16}, 
xticklabels={1,2,3,4,5,6,7,8,9,10,11,12,13,14,15,16},
every major tick/.style={black, semithick},
y grid style={lightgray!92.02614379084967!black},
ylabel={accuracy [\%]},
ymajorgrids,
ymin=85, 
ymax=100,
yminorgrids,
minor y tick num=2
]

\addplot [very thick, color1, mark=triangle, mark size=2, mark options={solid}]
table [row sep=\\ ]{%
1	98.32	\\
2	98.62	\\
3	98.98	\\
4	99.18	\\
5	99.17	\\
6	99.27	\\
7	99.28	\\
8	99.37	\\
9	99.28	\\
10	99.38	\\
11	99.35	\\
12	99.47	\\
13	99.48	\\
14	99.43	\\
15	99.47	\\
16	99.48	\\
};
\addplot [very thick, color3, mark=diamond, mark size=2, mark options={solid}]
table [row sep=\\ ]{%
1	98.22	\\
2	98.95	\\
3	98.97	\\
4	99.02	\\
5	99.03	\\
6	99.27	\\
7	99.3	\\
8	99.42	\\
9	99.2	\\
10	99.25	\\
11	99.4	\\
12	99.35	\\
13	99.28	\\
14	99.3	\\
15	99.27	\\
16	99.23	\\
};
\addplot [very thick, color4, mark=pentagon, mark size=2, mark options={solid}]
table [row sep=\\ ]{%
1	85.9	\\
2	91.42	\\
3	93.78	\\
4	94.92	\\
5	95.1	\\
6	95.15	\\
7	95.72	\\
8	96.12	\\
9	95.83	\\
10	96.42	\\
11	96.45	\\
12	96.35	\\
13	96.25	\\
14	96.42	\\
15	96.43	\\
16	96.4	\\
};
\addplot [very thick, color5, mark=o, mark size=2, mark options={solid}]
table [row sep=\\ ]{%
1	86.9	\\
2	89.55	\\
3	91.15	\\
4	90.97	\\
5	91.52	\\
6	91.43	\\
7	91.4	\\
8	91.32	\\
9	91.2	\\
10	92	\\
11	92.3	\\
12	92.3	\\
13	92.33	\\
14	92.22	\\
15	92.37	\\
16	92.38	\\
};
\addplot [very thick, color7, mark=square, mark size=2, mark options={solid}]
table [row sep=\\ ]{%
1	95.72	\\
2	98.08	\\
3	98.83	\\
4	99.12	\\
5	99.1	\\
6	99.22	\\
7	99.17	\\
8	99.27	\\
9	99.28	\\
10	99.48	\\
11	99.35	\\
12	99.38	\\
13	99.42	\\
14	99.48	\\
15	99.45	\\
16	99.42	\\
};
\end{axis}
\end{tikzpicture}
	\caption{Accuracy scores for $112\px$ resolution testing on the LFW dataset of the baseline (diamonds), \BT-56 (triangles), \BT-7 (pentagons) and \ST-56 (squares), and \ST-7 (circles) after each training epoch.}
	\label{fig:learning_curve}
\end{figure}
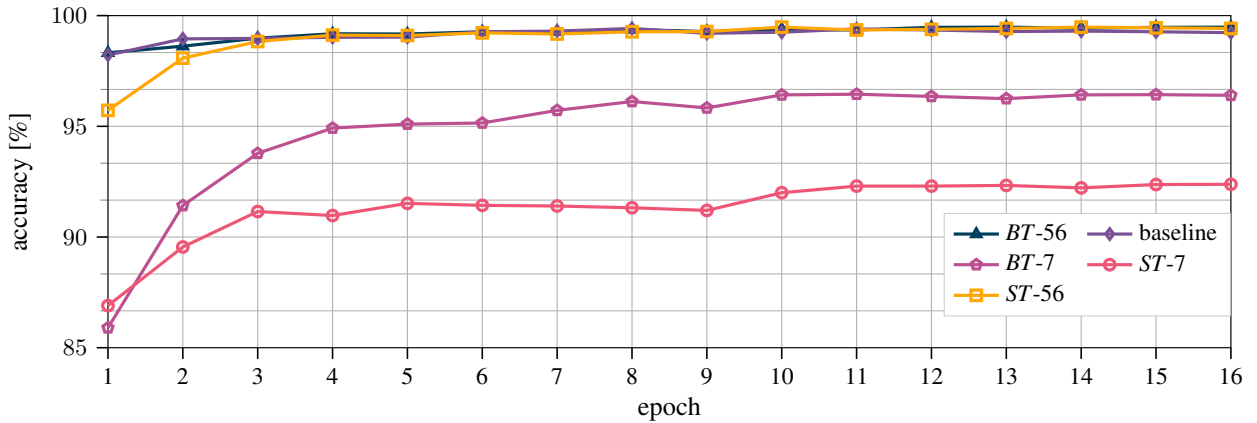

Finishing this chapter, a comparison between all introduced methods is given. First, we analyze the verification performance on HR images for both proposed methods and compare them to the baseline approach. \Cref{fig:learning_curve} shows the accuracy of the LFW dataset for each epoch. We select the models \BT-7, \BT-56 and \ST-7, \ST-56 to represent a very low and relatively high resolution. After the first epoch our baseline model achieves about $98\%$ accuracy, followed by almost peak accuracy already after the second epoch. During epochs $3$ and $16$, no significant changes in accuracy are visible. The \BT-56 starts with equal accuracy after the first epoch and then takes another two epochs to reach almost peak accuracy. The \ST-56 gets only after epoch $4$ approximately peak performance. This model needs significantly more samples than both previously mentioned models to achieve similar accuracy. One reason for that could be the additional feature distance loss, which forces the network to additionally concentrate on minimizing feature distances and not only learn a good classification. 

The peak performance for both methods \BT-7 and \ST-7 is significantly lower compared to the other approaches. This could evolve from too little information in the very low LR images, which might probably be just too less resolution to be able to learn a proper feature extraction. Moreover, the speed of learning is clearly slower. To get close to the maximum accuracy, both approaches need at least $10$ epochs. 

Second, table~\ref{table:1} compares all presented methods towards their training time per epoch and performance in the two-resolution scenario and multi-resolution scenario and depicts accuracy values on the LFW dataset. We conduct that compared to the two resolution techniques, both \ST-M1 and \ST-M2 models clearly outperform the baseline and the \BT-M for low resolutions. Focusing on higher resolutions, the \BT-M method is performing best. Even for the original image resolution of $112\px$, the \BT-M model performs better than the baseline. We think this is reasonable because using lower resolutions additionally during training can be seen as an extra data augmentation and hence, can improve the performance. One also has to compromise that for a performance improvement of about absolute $14\%$ in the low\_res protocol, the performance for high\_res drops about $2\%$. The second siamese training approach, \ST-M2 is performing worse in all categories than \ST-M1. Therefore, we conclude that the much greater effort for training is not worth it. It seems to be less essential to force a network to learn close features for the same image in different resolutions, within each batch, than across several batches. 

\begin{table}[t]
    \centering
    \caption{Comparison in terms of training time per epoch and accuracy on LFW dataset for different resolution protocols. Bold numbers denote the best performance across all methods.}
    
\begin{tabular}{clrrrrrr}
  \toprule
      &       & \multicolumn{1}{c}{baseline} & \multicolumn{1}{c}{\BT-5} & \multicolumn{1}{c}{\BT-M} & \multicolumn{1}{c}{\ST-5} & \multicolumn{1}{c}{\ST-M1} & \multicolumn{1}{c}{\ST-M2} \\
  \midrule
      training time &  & \multicolumn{1}{c}{$2\,\mathrm{h}$} & \multicolumn{1}{c}{$2\,\mathrm{h}$} & \multicolumn{1}{c}{$2\,\mathrm{h}$}  & \multicolumn{1}{c}{$4\,\mathrm{h}$} & \multicolumn{1}{c}{$4\,\mathrm{h}$} & \multicolumn{1}{c}{$20\,\mathrm{h}$} \\
\multirow{6}[0]{*}{accuracy } & $112\px$ & $99.23\%$ & \multicolumn{1}{c}{\multirow{5}[0]{*}{.}} & $\textbf{99.30\%}$ & \multicolumn{1}{c}{\multirow{5}[0]{*}{.}} & 97.40\% & $95.62\%$ \\
      & all\_res & $96.86\%$ &       & $\textbf{97.72\%}$ &       & $96.76\%$ & $95.07\%$ \\
      & high\_res & $99.20\%$ &       & $\textbf{99.33\%}$ &       & $97.35\%$ & $95.62\%$ \\
      & mid\_res & $95.89\%$ &       & $\textbf{97.78\%}$ &       & $96.98\%$ & $95.51\%$ \\
      & low\_res & $77.57\%$ &       & $87.17\%$ &       & $\textbf{91.50\%}$ & $88.72\%$ \\
      & $5\px$   & $54.65\%$ & $69.66\%$ & $71.53\%$ & $67.86\%$ & $\textbf{76.78\%}$ & $71.84\%$ \\
      \bottomrule
\end{tabular}
    \label{table:1}
\end{table}

Lastly, the inference time and the number of parameters are equal for all models, which make the comparison fair and reasonable. 

\section{Conclusions and Future Work}
\label{sec:conclusion}
In this work, we analyze the impact of different image resolutions on face verification performance using a state-of-the-art approach. The distances between extracted features are deeply analyzed. Our findings are that features of classical face recognition networks are not scale-invariant. The performance decreases heavily for lower resolutions.

To achieve best performance, the image resolution requires to be the same as in the corresponding training dataset for the network. To overcome this problem, we propose two intuitive methods to learn scale-invariant features directly. 1) \BT describes training our network with batches containing LR and HR images in the same ratio. 2) \ST is a siamese network structure constructed from a state-of-the-art network. This approach is moreover using feature-distances between LR and HR version of the same image as an additional loss to minimize these distances besides optimizing the classification problem. We demonstrate that our models can clearly improve CR face verification accuracy on five standard datasets. 

Moreover, we train our proposed models with several resolutions at once. Hence, a single model can be applied to arbitrary image scales. We also report considerable improvements for CR verification performance, especially for low resolutions. 

Lastly, we introduce three different evaluation protocols for five popular datasets, considering multiple resolutions in the face verification evaluation process for CR scenarios.

In the future, we plan to focus more on minimizing the intra-class feature-distance between LR and HR images to improve the robustness of features furthermore. Moreover, the inter-class distances between LR and HR images should be further maximized. Therefore, we consider adding and adapting a triplet loss function to our networks. We also want to extend the downsample process by using arbitrary blur kernels like described in~\cite{zhang2019deep} and apply it in our work.

\bibliographystyle{IEEEbib}
\bibliography{main}

\end{document}